\documentclass[sigconf]{acmart}
\AtBeginDocument{%
  \providecommand\BibTeX{{%
    \normalfont B\kern-0.5em{\scshape i\kern-0.25em b}\kern-0.8em\TeX}}}

\setcopyright{none}
\makeatletter
\renewcommand\@formatdoi[1]{\ignorespaces}
\makeatother
\renewcommand{\footnotetextcopyrightpermission}[1]{}
\acmSubmissionID{1935}

\usepackage{bm}
\usepackage{subfigure}
\usepackage{multirow}
\usepackage{array}

\begin{document}

\title{Diverse Semantics-Guided Feature Alignment and Decoupling for Visible-Infrared Person Re-Identification}
\author{
    Neng Dong\textsuperscript{\rm 1}, Shuanglin Yan\textsuperscript{\rm 1}, Liyan Zhang\textsuperscript{\rm 2}, Jinhui Tang\textsuperscript{\rm 1}
}
\affiliation{
    \institution{\textsuperscript{\rm 1}Nanjing University of Science and Technology\\
    \textsuperscript{\rm 2}Nanjing University of Aeronautics and Astronautics\\
    \textsuperscript{\rm }}
    \country{}
}





  


\renewcommand{\shortauthors}{Neng Dong, et al.}

\begin{abstract}
Visible-Infrared Person Re-Identification (VI-ReID) is a challenging task due to the large modality discrepancy between visible and infrared images, which complicates the alignment of their features into a suitable common space. Moreover, style noise, such as illumination and color contrast, reduces the identity discriminability and modality invariance of features. To address these challenges, we propose a novel Diverse Semantics-guided Feature Alignment and Decoupling (DSFAD) network to align identity-relevant features from different modalities into a textual embedding space and disentangle identity-irrelevant features within each modality. Specifically, we develop a Diverse Semantics-guided Feature Alignment (DSFA) module, which generates pedestrian descriptions with diverse sentence structures to guide the cross-modality alignment of visual features. Furthermore, to filter out style information, we propose a Semantic Margin-guided Feature Decoupling (SMFD) module, which decomposes visual features into pedestrian-related and style-related components, and then constrains the similarity between the former and the textual embeddings to be at least a margin higher than that between the latter and the textual embeddings. Additionally, to prevent the loss of pedestrian semantics during feature decoupling, we design a Semantic Consistency-guided Feature Restitution (SCFR) module, which further excavates useful information for identification from the style-related features and restores it back into the pedestrian-related features, and then constrains the similarity between the features after restitution and the textual embeddings to be consistent with that between the features before decoupling and the textual embeddings. Extensive experiments on three VI-ReID datasets demonstrate the superiority of our DSFAD.
\end{abstract}


\keywords{Visible-Infrared Person Re-Identification, Diverse Semantics, Feature Alignment, Feature Decoupling, Feature Restitution.}

\maketitle


\section{Introduction}
Person Re-identification (ReID) \cite{mvip, LCR2S, tal, shen2023pbsl, propot} aims to retrieve pedestrian images with the same identity across different scenes, playing a significant role in intelligent surveillance and security monitoring \cite{m3net, trimatch, lin2022learning, wang2022body, tang2024divide}. However, this task primarily focuses on matching visible images captured during the daytime, making it unsuitable for nighttime scenarios. To address this limitation, Visible-Infrared Person Re-Identification (VI-ReID) \cite{vireid} has been proposed, enabling cross-modality retrieval between visible and infrared images. It satisfies the requirements of intelligent surveillance for continuous, round-the-clock monitoring, offering substantial application values and research implications.
\begin{figure}[t!]
	\centering
        \includegraphics[width=3.3in,height=2.0in]{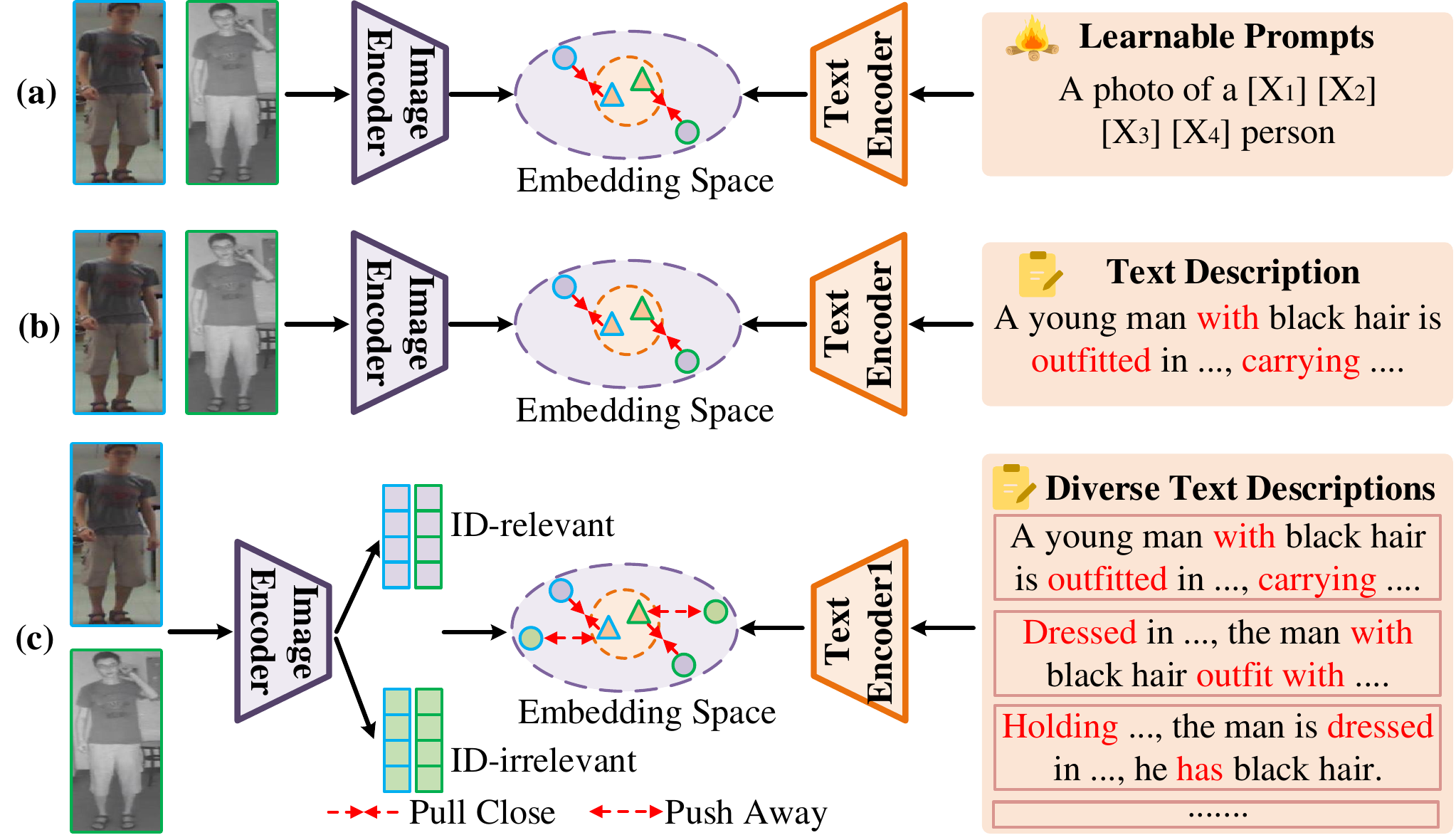}
	\caption{The core motivation and main idea of our DSFAD. (a) Learnable textual prompts lack clear and detailed pedestrian semantics. (b) The description generated directly with LLaVA follows a fixed sentence structure. (c) Our descriptions with diverse sentence structures prevent the model from overfitting to rigid semantic patterns. Additionally, our method effectively disentangles identity-irrelevant information.}
	\label{Fig:1}
\end{figure}

In contrast to the extensively studied single-modality person ReID \cite{bagtricks, pcb, entdnet, cajd}, VI-ReID poses greater challenges due to the substantial imaging discrepancies between visible and infrared images. To address this issue, both image-level and feature-level modality alignment methods have been proposed. Image-level methods \cite{aligngan, hicmd, d2rl} transfer the style of images to another modality or an intermediate modality. However, these approaches rely on additional generative networks, and the synthesized images often suffer from distortions, which impede the learning of discriminative pedestrian representations. Feature-level methods \cite{nfs, maum, deen, dpli} aim to optimize the network structure and metric function to project visual features from different modalities into a common embedding space. While effective, these methods face two primary issues: first, the learned visual features lack high-level semantic information; second, the large modality discrepancies make it difficult to align heterogeneous visual features into a suitable common space.

Large Language-Vision Models (LLVMs), such as CLIP \cite{clip}, have proven highly effective in addressing the above challenges. First, matching image-text pairs enables the VI-ReID models to capture pedestrian semantics. Second, aligning visual features from different modalities into the textual embedding space is more efficient, as language descriptions corresponding to heterogeneous images exhibit no modality discrepancy. Notably, pedestrian images typically lack textual descriptions, which poses difficulties in directly applying CLIP to VI-ReID. Figure \ref{Fig:1} illustrates two feasible approaches. The first \cite{csdn} incorporates CoOp \cite{coop} to train learnable textual prompts. However, they struggle to capture explicit pedestrian semantics and this strategy typically requires multi-stage training. The second approach \cite{eees} employs LLaVA \cite{llava} to generate the specific textual description of pedestrian appearances. While it provides clearer and more detailed semantics than the former, we observe that the generated descriptions for all images follow a rigid sentence structure, leading to the model overfitting to non-differentiated semantic patterns. Furthermore, most existing methods neglect the filtering of style information, such as illumination and color contrast, which vary significantly across different images of the same pedestrian. These variations are particularly pronounced between visible and infrared images, as the former is captured during the daytime and the latter at nighttime, thereby reducing the identity discriminability and modality invariance of visual features.

To address the above challenges, we propose a novel Diverse Semantics-guided Feature Alignment and Decoupling (DSFAD) network to align the identity-relevant visual features between modalities and to filter out the identity-irrelevant ones within each modality. It consists of Diverse Semantics-guided Feature Alignment (DSFA), Semantic Margin-guided Feature Decoupling (SMFD), and Semantic Consistency-guided Feature Restitution (SCFR) modules. In DSFA, we interact with ChatGPT \cite{chatgpt} to obtain varied sentence templates for describing pedestrian appearances. For each visible and infrared image, we then randomly select a template to request LLaVA to generate the corresponding textual description. We maximize the similarity between images and texts, ensuring that the visual features are rich in pedestrian semantics while aligning them from both modalities into the textual embedding space. To disentangle style information that is independent of pedestrian semantics, SMFD decomposes the visual features into pedestrian-relevant and pedestrian-irrelevant components, and constrains the similarity between the former and textual embeddings to be at least a margin higher than that between the latter and textual embeddings. Considering the potential loss of pedestrian semantics during feature decoupling, SCFR further excavates the information that positively contributes to identification from the style-related features, restores it back into the pedestrian-related features, and constrains the similarity between features after restitution and textual embeddings to be consistent with that between features before decoupling and textual embeddings. Our DSFAD is single-stage and trained end-to-end. During inference, textual descriptions are not required and only identity-irrelevant features are extracted, simple and efficient.

Our main contributions are summarized as follows:
\begin{itemize}
\item We propose DSFA to learn semantically-enriched visual features, to align visible and infrared features into the textual embedding space, and to reduce the risk of the model overfitting to non-differentiated semantic patterns.
\item We develop SMFD to disentangle style-related features, enabling the model to effectively filter out identity-irrelevant information. To the best of our knowledge, we are the first to explicitly address this issue in VI-ReID.
\item We design SCFR to restore the identity-relevant features, effectively preventing the loss of pedestrian-related semantics during feature decoupling.
\item Extensive experiments on three benchmark datasets validate the superiority of our proposed method and demonstrate the effectiveness of each designed module.
\end{itemize}

\vspace{-0.21cm}
\section{Related Work}

\subsection{VI-ReID based on Image-level Alignment}
Image-level modality alignment methods mitigate the discrepancies between visible and infrared images through style transfer between the two modalities. For example, Wang \emph{et al.} \cite{d2rl} introduced a generative adversarial network \cite{gan} to transform infrared images into their visible counterparts and vice versa. To prevent the loss of identity information during the style transferring, Choi \emph{et al.} \cite{hicmd} designed an attribute-guided person image generation strategy to retain the discriminative cues present in the original images. Furthermore, to address the ill-posed infrared-to-visible translation, Li \emph{et al.} \cite{xiv-reid} developed an auxiliary \emph{X} modality that maintains both the appearance of visible images and the shape information of infrared images. Recently, benefiting from the powerful creation capability of diffusion models, Pan \emph{et al.} \cite{acd} adopted a diffusion-based generator to achieve more efficient style transfer. However, these methods are often susceptible to mode collapse and tend to introduce noise, making it difficult to learn discriminative pedestrian representations. In contrast, our DSFAD avoids these issues and does not require the parameter-heavy generative network.

\subsection{VI-ReID based on Feature-level Alignment}
Feature-level modality alignment methods have recently gained popularity, focusing on designing effective feature extractors and metric functions to project visible and infrared features into a shared embedding space. Specifically, Chen \emph{et al.} \cite{nfs} proposed a cross-modality contrastive optimization criterion to minimize the distance between visible and infrared embeddings of the same identity. To learn features that are both modality-invariant and identity-discriminative, Wu \emph{et al.} \cite{mpanet} developed a modality alleviation module to dislodge style discrepancies and a pattern alignment module to discover pedestrian nuances. Moreover, Liu \emph{et al.} \cite{maum} introduced a modality-agnostic classifier and two unidirectional metrics to guide the alignment of feature prototypes across visible and infrared modalities. To learn more informative representations, Zhang \emph{et al.} \cite{deen} adopted multi-branch convolutional blocks that extract diverse features for training. Despite these advances, they lack the ability to capture high-level semantic information, and the large modality discrepancy still poses challenges in aligning features into a suitable common space. In contrast, our DSFAD introduces textual modality to guide the learning and alignment of visual features, addressing these limitations effectively.

\begin{figure*}[t!]
	\centering
	\includegraphics[width=6.8in,height=3.2in]{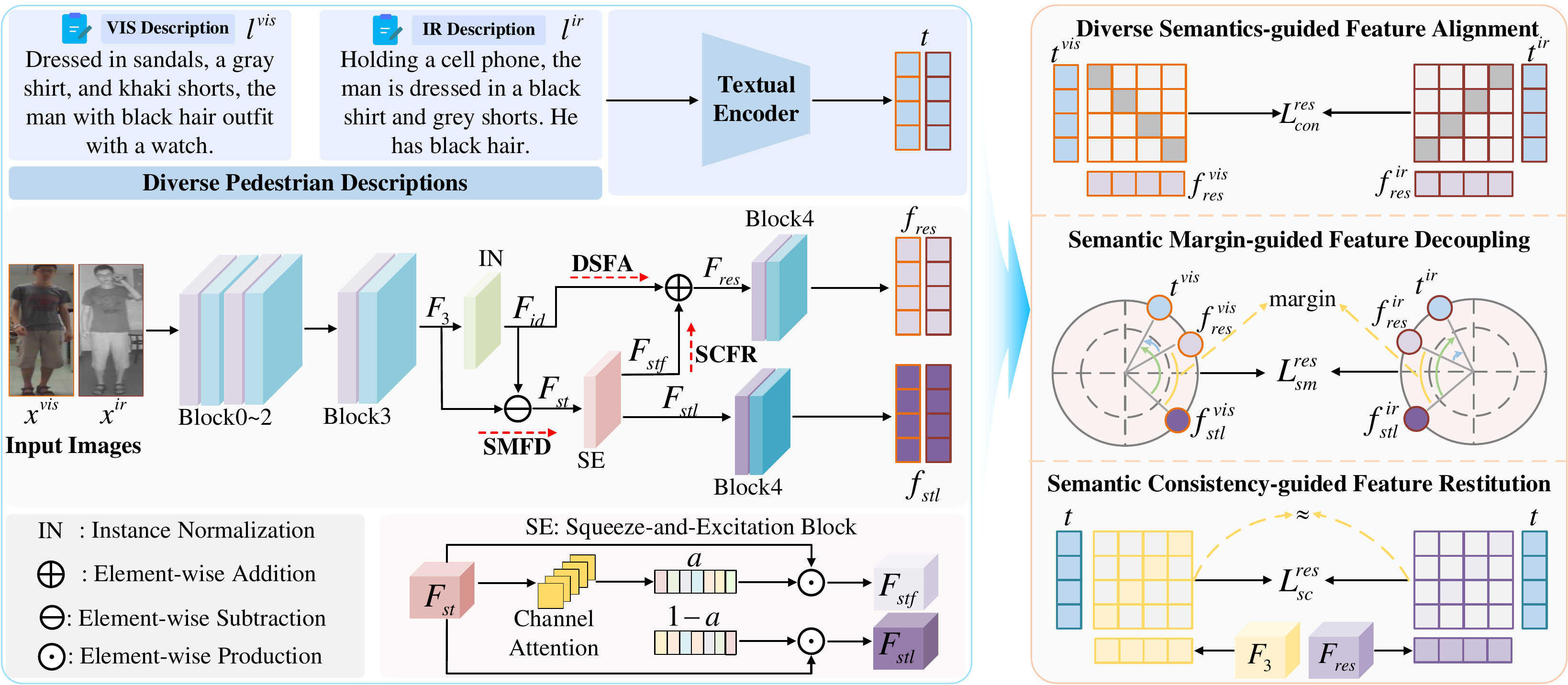}
	\caption{The overview of our proposed DSFAD. It consists of DSFA, SMFD, and SCFR. DSFA employs pedestrian descriptions with diverse sentence structures to guide the alignment of cross-modality identity-relevant features. SMFD decouples within-modality identity-irrelevant features through the semantic margin constraint. SCFR prevents the loss of pedestrian semantics through the semantic consistency constraint. During inference, only the visual component of DPSFA is required.}
	\label{Fig:2}
\end{figure*}

\subsection{Large Language-Vision Models for ReID}
Large language-vision models, particularly CLIP, have exhibited remarkable performance across various downstream vision tasks \cite{detpro, cris, winclip, shen2024imagpose, shen2025imagdressing} due to their exceptional capability in capturing image semantics. In the realm of person ReID, Li \emph{et al.} \cite{clipreid} formulated the first CLIP-ReID paradigm that incorporates CLIP and CoOP to drive the model to learn visual features with pedestrian semantics. Yan \emph{et al.} \cite{cfine} proposed a CLIP-driven fine-grained information excavation network to transfer rich prior knowledge from CLIP to text-image person ReID. Recently, to alleviate the modality gap in VI-ReID, Yu \emph{et al.} \cite{csdn} introduced learnable textual prompts for visible and infrared images, aligning heterogeneous visual features into the textual embedding space. However, the pedestrian semantics learned through these prompts are unknown and lack fine-grained detail. To address this issue, Dong \emph{et al.} \cite{eees} employed LLaVA to generate specific pedestrian descriptions. However, the rigid sentence structures led the VI-ReID model to overfit the repetitive semantic patterns. Additionally, existing VI-ReID methods struggle to filter out identity-irrelevant style information, which weakens the discriminability of features. In contrast, our DSFAD effectively guides the alignment of inter-modality identity-relevant features and the separation of intra-modality identity-irrelevant features.

\section{The proposed method}

The overview of our DSFAD is illustrated in Figure \ref{Fig:2}. Section \ref{sec:3.1} introduces DSFA, which aligns visible and infrared features into the semantic space with the guidance of pedestrian descriptions featuring diverse sentence structures. Section \ref{sec:3.2} presents SMFD, which addresses the challenge of accurately filtering out style information. Section \ref{sec:3.3} discusses the SCFR, designed to prevent the loss of pedestrian semantics during feature decoupling. Section \ref{sec:3.4} summarizes the processes of training and inference.

\subsection{Diverse Semantics-guided Feature Alignment}
\label{sec:3.1}
Applying CLIP to facilitate VI-ReID has recently gained popularity \cite{csdn, eees}, not only due to its ability to assist visual features in sensing high-level semantic information but, more importantly, because aligning visible and infrared visual features into the textual embedding space effectively alleviates the modality discrepancy. However, a key challenge in applying CLIP to VI-ReID is the absence of textual descriptions for pedestrian images. To address this, Yu \emph{et al.} \cite{csdn} designed the trainable textual prompts `\emph{A photo of a [$X_{1}$] [$X_{2}$] [$X_{3}$] [$X_{4}$] person.}' to implicitly learn pedestrian semantics. Although effective, these prompts fail to adequately characterize pedestrians as their semantics are unknown and coarseness. Recently, inspired by the impressive capability in image captioning of LLaVA, Dong \emph{et al.} \cite{eees} generates the specific textual description such as `\emph{A young man with dark hair is outfitted in a gray shirt and khaki shorts, accompanied by a watch.}' to provide explicit pedestrian semantics. While this approach alleviates the limitation of learnable prompts, we observe that the generated descriptions for all images consistently adhere to a fixed sentence structure `\emph{A [age group] [gender] with [hair color] [hairstyle] is outfitted in [upper clothing] and [lower clothing], accompanied by [accessories].}', causing the VI-ReID model to overfit to rigid semantic patterns. To this end, we propose DSFA to generate pedestrian descriptions with varied sentence structures, guiding more efficient semantic sensing and modality alignment.

\begin{figure}[t!]
	\centering
        \includegraphics[width=3.3in,height=3.9in]{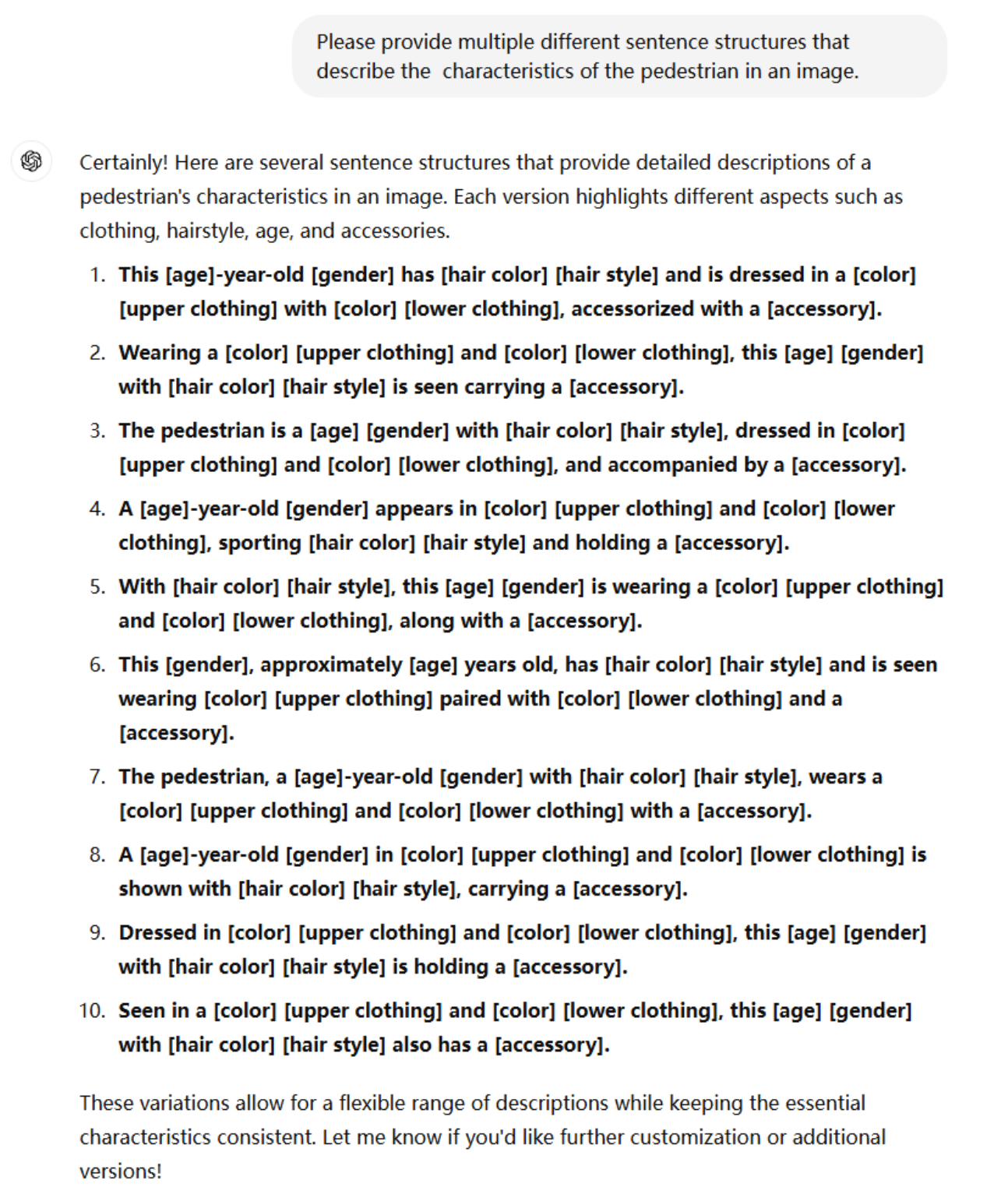}
	\caption{The process of acquiring description templates featuring diverse sentence structures with ChatGPT.}
	\label{Fig:3}
\end{figure}

Specifically, as illustrated in Figure \ref{Fig:3}, we first interact with ChatGPT using the command `\emph{Please provide multiple different sentence structures that describe the characteristics of the pedestrian in an image.}'. It responds with 10 distinct description templates that highlight different aspects such as clothing, hairstyle, age, and accessories. Next, when given a pedestrian image, we randomly select a template `$T_{p}$' from ChatGPT's response and send the instruction `\emph{Please according to the template `$T_{p}$' to describe the characteristics of the pedestrian in the image.}' to LLaVA. In this way, descriptions we generated for different images with the same identity within each mini-batch may vary, significantly enhancing the diversity of pedestrian semantics. Suppose that the generated textual descriptions for cross-modality images $x=(x^{vis},x^{ir})$ are $l=(l^{vis},l^{ir})$, we utilize CLIP's visual encoder and textual encoder to extract their feature vectors $f=(f^{vis},f^{ir})$ and $t=(t^{vis},t^{ir})$, respectively. After that, we employ contrastive loss to maximize the similarity between them, ensuring the semantic relevance and modality invariance of visual features. This process can be formulated as:
\begin{equation}
\label{Eq:(1)}
\begin{aligned}
     L_{con}=&-\frac{1}{2PK}\sum_{i=1}^{2PK}\log\frac{\exp(s(f_{i},t_{i}))}{\sum\nolimits_{j=1}^{2PK}\exp(s(f_{i},t_{j}))} \\
     &-\frac{1}{2PK}\sum_{i=1}^{2PK}\log\frac{\exp(s(t_{i},f_{i}))}{\sum\nolimits_{j=1}^{2PK}\exp(s(t_{i},f_{j}))},
\end{aligned}
\end{equation}
where $P$ and $K$ represent the number of pedestrians sampled per modality and the number of images sampled for each pedestrian in each batch. $s(\cdot)$ denotes the cosine similarity of two feature vectors.

\subsection{Semantic Margin-guided Feature Decoupling}
\label{sec:3.2}
In addition to identity cues, pedestrian images often contain some environment-related style information, such as illumination and color contrast, which act as noise that increases the difficulty of identity matching. Moreover, due to variations in the environment and time of image capture in real-world scenarios, different images of the same pedestrian typically exhibit significant illumination and color discrepancies. These differences are particularly pronounced in visible and infrared images, with the former captured during the daytime under bright lighting conditions and the latter taken at nighttime with low-light exposure, significantly impacting the effectiveness of cross-modality visual features alignment. To address these challenges, we propose SMFD to separate the identity-relevant pedestrian features and the identity-irrelevant style features.

Specifically, as illustrated in Figure \ref{Fig:2}, we first incorporate an Instance Normalization (IN) \cite{in} layer following Block3 in the visual encoder, which has been demonstrated to effectively remove instance-irrelevant information \cite{snr, manet}. Let features output from Block3 be denoted as $F_{3}=(F_{3}^{vis}, F_{3}^{ir})$, the identity-relevant pedestrian features retained after applying IN can be represented as:
\begin{equation}
    F_{id} =IN(F_{3})=\gamma(\frac{F_{3}-\mu(F_{3})}{\sigma(F_{3})})+\beta,
\end{equation}
where $\mu(\cdot)$ and $\sigma(\cdot)$ denote the mean and standard deviation across spatial dimensions for each channel and each image, $\gamma(\cdot)$ and $\beta(\cdot)$ are parameters learned from the data. Next, we subtract $F_{id}$ from $F_{3}$ to obtain the remaining features $F_{st}=(F_{st}^{vis},F_{st}^{ir})$ encompassing identity-independent style information, formulated as:
\begin{equation}
    F_{st} =F_{3}-F_{id}.
\end{equation}

We then feed $F_{id}$ and $F_{st}$ into two Block4 layers with the same architecture but no shared parameters, followed by average pooling, obtaining the pedestrian-related and style-related feature vectors $f_{id}=(f^{vis}_{id}, f^{ir}_{id})$ and $f_{st}=(f^{vis}_{st}, f^{ir}_{st})$. Compared with the directly extracted feature vectors $f$, $f_{id}$ can promote more effective identity matching and modality alignment. However, despite the model possessing basic feature decoupling capabilities after IN processing, identity-related features and style-related features inevitably capture overlapping representational patterns as they are derived from the same input, hindering complete information separation. To address this issue, we further design a semantics-guided margin loss, which constrains the similarity between $f_{id}$ and $t$ to be at least a margin higher than that between $f_{st}$ and $t$:
\begin{equation}
\label{Eq:(4)}
     L_{sm}=\frac{1}{2PK}\sum_{i=1}^{2PK}max(0,m+s(f_{st,i},t_{i})-s(f_{id,i},t_{i})),
\end{equation}
where $m$ is the margin value and we empirically set it to 1.0. By minimizing $L_{sm}$, in the textual embedding space, $f_{id}$ is brought close to $t$ while $f_{st}$ is pushed away from $t$. This ensures that $f_{id}$ is related to pedestrian semantics and $f_{st}$ is independent of them, achieving a more effective decoupling of identity and style information.

\subsection{Semantic Consistency-guided Feature Restitution}
\label{sec:3.3}
Notably, the removed style-related features may contain information that contributes positively to identification. This arises primarily because IN may inadvertently discard essential pedestrian semantics \cite{snr, manet} while eliminating instance-irrelevant information, thereby weakening the discriminability of the pedestrian-related features. To prevent this potential loss, we further propose SCFR to compensate the pedestrian semantic-related information in the style-related features back to the pedestrian-related features.

Specifically, to selectively recover identification-useful information from the style-related features, we first introduce Squeeze-and-Excitation Networks (SE) \cite{se} after the IN layer. As illustrated in Figure \ref{Fig:2}, the SE module employs a channel attention mechanism that adaptively learns a set of channel-wise weights, denoted as $a=[a_{1}, a_{2},\cdots, a_{c}]$, where $c$ represent the number of channels in $F_{st}$. These weights dynamically adjust the relative importance of each channel in $F_{st}$, which can be formulated as:
\begin{equation}
     a=\sigma(W_{2}\delta(W_{1}(pool(F_{st})))),
\end{equation}
where $\sigma$, $\delta$, and $pool$ represent the Sigmoid activation function, the ReLU activation function, and the average pooling operation respectively. $W_{1}$ and $W_{2}$ are two fully connected layers. 

With the learnable $a$, we further disentangle $F_{st}$ into identification-useful component $F_{stf}$ and identification-useless component $F_{stl}$:
\begin{equation}
     F_{stf}=a \odot F_{st},
\end{equation}
\begin{equation}
     F_{stl}=(1-a) \odot F_{st},
\end{equation}
where $\odot$ indicates the element-wise product operation, ensuring that each channel is weighted accordingly. 

We then add $F_{stf}$ to the pedestrian-related features $F_{id}$ to reintroduce identification-useful information discarded during style removal. The obtained restored features can be represented as:
\begin{equation}
     F_{res}=F_{id} + F_{stf}.
\end{equation}

Ideally, the features after restitution should enjoy as much pedestrian semantic information as the original features before style removal, which means that the information discarded by IN has been effectively recovered. To this end, we develop a semantic-guided consistency loss, which enforces that the similarity between $F_{res}$ and $t$ remains consistent with the similarity between $F_3$ and $t$:
\begin{equation}
     L_{sc}^{res}=\frac{1}{2PK}\sum_{i=1}^{2PK}(s(pool(F_{3,i}),t_{i})-s(pool(F_{res,i}),t_{i})).
\end{equation}

This constraint quantifies the semantic relevance of $F_{res}$ and $F_{3}$ by comparing their similarity to $t$. By minimizing it, we ensure $F_{res}$ remains semantically aligned with $F_{3}$, thereby mitigating the negative effects of IN and enhancing the discriminability of features.

\subsection{Training and Inference}
\label{sec:3.4}
After the aforementioned feature decoupling and restitution, as illustrated in Figure \ref{Fig:2}, the resulting cross-modality visual feature vectors that are fully identity-relevant and entirely identity-irrelevant are denoted as $f_{res}=(f_{res}^{vis},f_{res}^{ir})$ and $f_{stl}=(f_{stl}^{vis},f_{stl}^{ir})$, respectively. Accordingly, we reformulate the contrastive loss (Eq. (\ref{Eq:(1)})) in the DSFA to align $f_{res}$ into the textual embedding space:
\begin{equation}
\begin{aligned}
     L_{con}^{res}=&-\frac{1}{2PK}\sum_{i=1}^{2PK}\log\frac{\exp(s(f_{res,i},t_{i}))}{\sum\nolimits_{j=1}^{2PK}\exp(s(f_{res,i},t_{j}))} \\
     &-\frac{1}{2PK}\sum_{i=1}^{2PK}\log\frac{\exp(s(t_{i},f_{res,i}))}{\sum\nolimits_{j=1}^{2PK}\exp(s(t_{i},f_{res,j}))}.
\end{aligned}
\end{equation}

Similarly, we rewrite the semantic-guided margin loss (Eq. (\ref{Eq:(4)})) in the SMFD to accurately separate $f_{res}$ and $f_{stl}$, ensuring $f_{res}$ is highly related to pedestrian semantics while $f_{stl}$ is completely independent of pedestrian semantics:
\begin{equation}
     L_{sm}^{res}=\frac{1}{2PK}\sum_{i=1}^{2PK}max(0,m+s(f_{stl,i},t_{i})-s(f_{res,i},t_{i})).
\end{equation}

Moreover, to ensure the identity discriminability of $f_{res}$, we apply the common identity loss to optimize the network:
\begin{equation}
    L_{id} =-\frac{1}{2PK}\sum_{i=1}^{2PK}q_{i}\log(p_{i}),
\end{equation}
where $q_{i}$ represents the one-hot vector of identity label $y_{i}$, and $p_{i}$ is the classification result of $f_{res,i}$.

In addition, we introduce the modality-shared enhancement loss \cite{mse} to force the distance between cross-modality positive pairs to be closer to the distance between intra-modality positive pairs:
\begin{equation}
    L_{mse} =\frac{1}{2PK}\sum_{p=1}^{P}\Big[\sum_{k=1}^{2K}(d_{k}^{intra}-d_{k}^{inter})^{2}\Big],
\end{equation}
where $d_{k}^{intra}$ represents the Euclidean distance of intra-modality positive embedding pairs, while $d_{k}^{inter}$ denotes that of inter-modality positive embedding pairs.

Finally, the total training loss can be formulated as:
\begin{equation}
    L_{total}=L_{id}+L_{mse}+\lambda_{1}L_{con}^{res}+\lambda_{2}L_{sm}^{res}+\lambda_{3}L_{sc}^{res},
\end{equation}
where $\lambda_{1}$, $\lambda_{2}$, and $\lambda_{3}$ are the hyperparameters that balance the relative importance of the loss terms. 

During inference, only the visual encoder of DPSFA is used to extract identity-relevant feature vectors $f_{res}$ for matching identity, and the other components are not required.

\begin{table*}[!ht]\small
\centering {\caption{Comparisons between our DSFAD and recent state-of-the-art methods on the SYSU-MM01 dataset. The best performance of existing methods is highlighted in blue, while our results are bolded. ‘-’ indicates that no performance is available.}\label{Tab:1}
\renewcommand\arraystretch{1.1}
\begin{tabular}{c|c|cccc|cccc||cccc|cccc}
\hline
 & & \multicolumn{8}{c||}{All-Search} & \multicolumn{8}{c}{Indoor-Search}\\
  
 \cline{3-18} Methods & Ref & \multicolumn{4}{c|}{Single-Shot} & \multicolumn{4}{c||}{Multi-Shot} & \multicolumn{4}{c|}{Single-Shot} & \multicolumn{4}{c}{Multi-Shot}\\

  \cline{3-18}
  
  & & R1 & R10 & R20 & mAP & R1 & R10 & R20  & mAP & R1 & R10 & R20 & mAP & R1 & R10 & R20 & mAP \\
\hline

  MMN \cite{mmn} & MM'21 & 70.6 & 96.2 & 99.0 & 66.9 & - & - & - & - & 76.2 & 97.2 & 99.3 & 79.6 & - & - & - & - \\
  
  FMCNet \cite{fmcnet} & CVPR'22 & 66.3 & - & - & 62.5 & 73.4 & - & - & 56.0 & 68.1 & - & - & 74.0 & 78.8 & - & - & 63.8 \\

  DART \cite{dart} & CVPR'22 & 68.7 & 96.3 & 98.9 & 66.2 & - & - & - & - & 72.5 & 97.8 & 99.4 & 78.1 & - & - & - & - \\

  DCLNet \cite{dclnet} & MM'22 & 70.7 & - & - & 65.18 & - & - & - & - & 73.5 & - & - & 76.8 & - & - & - & - \\

  MAUM \cite{maum} & CVPR'22 & 71.6 & - & - & 68.7 & - & - & - & - & 76.9 & - & - & 81.9 & - & - & - & - \\

   CMT \cite{cmt} & ECCV'22 & 71.8 & 96.4 & 98.8 & 68.5 & 80.2 & 97.9 & 99.5 & 63.1 & 76.9 & 97.6 & 99.6 & 79.9 & 84.8 & 99.4 & 99.9 & 74.1 \\

    MSCLNet \cite{msclnet} & ECCV'22 & 76.9 & 97.6 & 99.1 & 71.6 & - & - & - & - & 78.4 & 99.3 & 99.9 & 81.1 & - & - & - & - \\
  
  PMT \cite{mse} & AAAI'23 & 67.5 & 95.3 & 98.6 & 64.9 & - & - & - & - & 71.6 & 96.7 & 99.2 & 76.5 & - & - & - & - \\

   CAJ+ \cite{caj+} & TPAMI'23 & 71.4 & 96.2 & 78.7 & 68.1 & - & - & - & - & 78.3 & 98.3 & 99.7 & 81.9 & - & - & - & - \\

   ProtoHPE \cite{protohpe} & MM'23 & 71.9 & 96.1 & 97.9 & 70.5 & - & - & - & - & 77.8 & 98.6 & 99.5 & 81.3 & - & - & - & - \\

    CAL \cite{cal} & ICCV'23 & 74.6 & 96.4 & - & 71.7 & 77.0 & 98.0 & - & 64.8 & 79.6 & 98.9 & - & 83.6 & 86.9 & 99.8 &  & 78.5 \\

   DEEN \cite{deen} & CVPR'23 & 74.7 & 97.6 & 99.2 & 71.8 & - & - & - & - & 80.3 & 99.0 & 99.8 & 83.3 & - & - & - & - \\

    MBCE \cite{mbce} & AAAI'23 & 74.7 & - & - & 72.0 & 78.3 & - & - & 65.7 & 83.4 & - & - & 86.0 & 88.4 & - & - & 80.6 \\

    SEFL \cite{sefl} & CVPR'23 & 75.1 & 96.8 & 99.1 & 70.1 & - & - & - & - & 78.4 & 97.4 & 98.9 & 81.2 & - & - & - & - \\

    MUN \cite{mun} & ICCV'23 & 76.2 & \textcolor{blue}{97.8} & - & 73.8 & - & - & - & - & 79.4 & 98.0 & - & 82.0 & - & - & - & - \\

      RLE \cite{rle} & NIPS'24 & 75.4 & 97.7 & \textcolor{blue}{99.6} & 72.4 & - & - & - & - & \textcolor{blue}{84.7} & \textcolor{blue}{99.3} & \textcolor{blue}{99.9} & \textcolor{blue}{87.0} & - & - & - & - \\

   HOS-Net \cite{hosnet} & AAAI'24 & 75.6 & - & - & \textcolor{blue}{74.2} & - & - & - & - & 84.2 & - & - & 86.7 & - & - & - & - \\

   CSCL \cite{cscl} & TMM'25 & 75.7 & - & - & 72.0 & - & - & - & - & 80.8 & - & - & 83.5 & - & - & - & - \\

    AGPI$^{2}$ \cite{agpi2} & TIFS'25 & 72.2 & 97.0 & - & 70.5 & - & - & - & - & 83.4 & 98.6 & - & 84.2 & - & - & - & - \\

     DPMF \cite{dpmf} & TNNLS'25 & 76.4 & 96.4 & 98.6 & 71.5 & - & - & - & - & 82.2 & 98.6 & 99.7 & 84.9 & - & - & - & - \\

   CSDN \cite{csdn} & TMM'25 & 76.7 & 97.2 & 99.1 & 73.0 & 83.5 & \textcolor{blue}{98.8} & \textcolor{blue}{99.7} & 67.9 & 84.5 & 99.2 & 99.6 & 86.8 & \textcolor{blue}{91.3} & \textcolor{blue}{99.8} & \textcolor{blue}{100.0} & \textcolor{blue}{82.2} \\

   ARGN \cite{argn} & TMM'25 & \textcolor{blue}{77.0} & - & - & 72.7 & \textcolor{blue}{84.3} & - & - & \textcolor{blue}{68.3} & 83.0 & - & - & 85.1 & 91.0 & - & - & 80.7 \\

  \hline
   \textbf{Ours (DSFAD)} & - & \textbf{78.4} & \textbf{97.5} & \textbf{99.3} & \textbf{74.7} & \textbf{84.6} & \textbf{98.5} & \textbf{99.6} & \textbf{69.9} & \textbf{85.1} & \textbf{99.2} & \textbf{99.7} & \textbf{87.2} & \textbf{91.5} & \textbf{99.7} & \textbf{99.9} & \textbf{82.7} \\
   \hline
\end{tabular}}
\end{table*}

\section{Experiments}
\subsection{Datasets and Experimental Settings}
\textbf{Datastes.} We evaluate the proposed DSFAD framework on three widely used VI-ReID datasets: SYSU-MM01 \cite{vireid}, RegDB \cite{regdb}, and LLCM \cite{deen}. SYSU-MM01 comprises 30,071 visible and 15,792 infrared images from 491 identities. Of these, 22,258 visible and 11,909 infrared images, corresponding to 395 identities, are used for training, while the remainder is reserved for testing. The dataset provides two search modes: all-search and indoor-search, as well as two gallery settings: single-shot and multi-shot. RegDB contains 4,120 visible and 4,120 infrared images from 412 identities, with 2,060 images per modality allocated for training and the rest for testing. It supports two evaluation strategies: visible-to-infrared and infrared-to-visible. LLCM consists of 25,626 visible and 21,141 infrared images from 1,064 identities. Among them, 16,946 visible and 13,975 infrared images are used for training, while the remaining images are designated for testing. The dataset offers two testing modes: visible-to-infrared and infrared-to-visible.

\textbf{Evaluation metrics.} We assess the retrieval performance with the standard Rank-$k$ accuracy and mean Average Precision (mAP).

\textbf{Implementation details.} We conduct experiments on a single NVIDIA GeForce 4090 GPU using the Pytorch framework. In each mini-batch, we randomly sample 32 images of 8 identities from each modality, with all images uniformly resized to 288$\times$144. Data augmentation techniques, including random padding, cropping, flipping, and channel exchange, are applied. We employ CLIP as the backbone, with ResNet50 \cite{resnet} as the image encoder and Transformer \cite{transformer} as the textual encoder. The initial learning rates for the image encoder and textual encoder are set to 3$\times$10$^{-4}$ and 1$\times$10$^{-6}$, respectively. The hyperparameters are set to $\lambda_{1}=0.15$, $\lambda_{2}=0.3$, and $\lambda_{3}=0.02$. We train the model using the Adam optimizer \cite{adam} for a total of 120 epochs, during which the initial learning rate is reduced to 0.1 times its value at the 40th epoch and is further decreased to 0.1 times the updated rate at the 70th epoch.

\begin{table}[!ht]\small
\centering {\caption{Comparisons on the RegDB dataset.}\label{Tab:2}
\renewcommand\arraystretch{1.1}
\begin{tabular}{c|ccc||ccc}
\hline
 \multirow{2}*{Methods} & \multicolumn{3}{c||}{Visible-to-Infrared} & \multicolumn{3}{c}{Infrared-to-Visible}\\

  & R1 & R10 & mAP & R1 & R10 & mAP \\
 \hline

 MMN & 91.6 & 97.7 & 84.1 & 87.5 & 96.0 & 80.5 \\

 FMCNet & 89.1 & - & 84.4 & 88.3 & - & 83.8 \\

 DART & 83.6 & - & 60.6 & 81.9 & - & 56.7 \\

 DCLNet & 81.2 & - & 74.3 & 78.0 & - & 70.6 \\

  MAUM & 87.8 & - & 85.0 & 86.9 & - & 84.3 \\

  CMT & 95.1 & 98.8 & 87.3 & 91.9 & 97.9 & 84.4 \\

  MSCLNet & 84.1 & - & 80.9 & 83.8 & - & 78.3 \\

  PMT & 84.8 & - & 76.5 & 84.1 & - & 75.1\\

   CAJ+ & 85.6 & 95.4 & 79.7 & 84.8 & 95.8 & 78.5 \\

   ProtoHPE & 88.7 & - & 83.7 & 88.6 & - & 81.9 \\

   CAL & 94.5 & \textcolor{blue}{99.7} & 88.6 & 93.6 & \textcolor{blue}{99.4} & 87.6 \\

   DEEN & 91.1 & 97.8 & 85.1 & 89.5 & 96.8 & 83.4 \\

   MBCE & 93.1 & 96.9 & 88.3 & 93.4 & 97.1 & 87.9\\

    SEFL & 91.0 & - & 85.2 & 92.1 & - & 86.5\\

    MUN & 95.1 & 98.9 & 87.1 & 91.8 & 97.9 & 85.0 \\

    RLE & 92.8 & 97.9 & 88.6 & 91.0 & 97.5 & 86.6 \\

    HOS-Net & 94.7 & - & \textcolor{blue}{90.4} & 93.3 & - & \textcolor{blue}{89.2} \\

    CSCL & 92.1 & - & 84.2 & 89.6 & - & 85.0 \\

   AGPI$^{2}$ & 89.0 & 98.1 & 83.8 & 87.9 & 97.1 & 83.0 \\

   DMPF & 88.8 & 97.3 & 81.0 & 88.8 & 97.6 & 81.8 \\

   CSDN & 95.4 & 98.8 & 87.7 & 92.3 & 98.0 & 85.5 \\

   ARGN & \textcolor{blue}{96.1} & - & 90.0 & \textcolor{blue}{94.1} & - & 87.8 \\

\hline
\textbf{Ours (DSFAD)} & \textbf{95.8} & \textbf{99.6} & \textbf{90.2} & \textbf{94.3} & \textbf{99.2} & \textbf{89.4} \\

  \hline
\end{tabular}}
\end{table}

\subsection{Comparison with State-of-the-art Methods}

\begin{table}[!ht]\small
\centering {\caption{Comparisons on the LLCM dataset.}\label{Tab:3}
\renewcommand\arraystretch{1.1}
\begin{tabular}{c|ccc||ccc}
\hline
 \multirow{2}*{Methods} & \multicolumn{3}{c||}{Visible-to-Infrared} & \multicolumn{3}{c}{Infrared-to-Visible}\\

  & R1 & R10 & mAP & R1 & R10 & mAP \\
 \hline

  MMN & 59.9 & - & 62.7 & 52.5 & - & 58.9 \\

  DART & 60.4 & 87.1 & 63.2 & 52.2 & 80.7 & 59.8 \\

 DEEN & 62.5 & 90.3 & 65.8 & 54.9 & 84.9 & 62.9 \\

 HOS-Net & \textcolor{blue}{64.9} & - & \textcolor{blue}{67.9} & 56.4 & - & 63.2 \\

 CSDN & 63.7 & \textcolor{blue}{90.9} & 66.5 & 55.8 & \textcolor{blue}{85.6} & \textcolor{blue}{63.5} \\

  ARGN & 63.9 & - & 66.6 & \textcolor{blue}{56.9} & - & 63.3 \\

\hline
\textbf{Ours (DSFAD)} & \textbf{66.2} & \textbf{91.5} & \textbf{68.9} & \textbf{57.5} & \textbf{86.2} & \textbf{64.1} \\

  \hline
\end{tabular}}
\end{table}

\textbf{Comparisons on SYSU-MM01.} We compare the proposed DSFAD with state-of-the-art methods on the SYSU-MM01 dataset, with results summarized in Table \ref{Tab:1}. Specifically, in the all-search and single-shot mode, DSFAD achieves a Rank-1 accuracy of 78.4\% and an mAP of 74.7\%, surpassing the previous best model, ARGN, by 1.4\% in Rank-1 accuracy and HOS-Net by 0.7\% in mAP. In the all-search and multi-shot mode, DSFAD outperforms ARGN by 0.3\% in Rank-1 accuracy and 1.6\% in mAP. For the indoor-search and single-shot mode, DSFAD improves Rank-1 accuracy from 84.7\% to 85.1\% and mAP from 87.0\% to 87.2\% compared to HOS-Net. Under the indoor-search and multi-shot mode, DSFAD surpasses CSDN by 0.2\% in Rank-1 accuracy and 0.5\% in mAP. These results demonstrate the superiority of DSFAD over existing methods on the SYSU-MM01 dataset across all evaluation settings.

\textbf{Comparisons on RegDB.} We also conduct comparisons between the proposed DSFAD and recent state-of-the-art methods on the RegDB dataset, as outlined in Table \ref{Tab:2}. DSFAD achieves superior recognition performance, attaining a Rank-1 accuracy of 95.8\% and a mAP of 90.2\% for visible-to-infrared retrieval, along with a Rank-1 accuracy of 94.3\% and an mAP of 89.4\% for infrared-to-visible retrieval. While DSFAD performs slightly worse than the best existing method in the visible-to-infrared setting, it surpasses state-of-the-art approaches in the infrared-to-visible setting. These results again confirm the competitive advantage of our DSFAD.

\textbf{Comparisons on LLCM.} We further evaluate our DSFAD on the LLCM dataset, with comparisons to state-of-the-art methods presented in Table \ref{Tab:3}. Specifically, in the visible-to-infrared retrieval setting, DSFAD surpasses HOS-Net by 1.3\% in Rank-1 accuracy and 1.0\% in mAP. In the infrared-to-visible retrieval setting, DSFAD outperforms ARGN by 0.6\% in Rank-1 accuracy and exceeds CSDN by 0.6\% in mAP. Furthermore, our DSFAD consistently achieves higher Rank-5 accuracy than CSDN across both retrieval settings. These results further confirm the superiority of the proposed DSFAD.

In summary, these comparisons clearly demonstrate the superiority of our proposed DSFAD, attributable to its ability to sense pedestrian-relevant semantics, to align visual features into the textual embedding space unrestricted by fixed semantic paradigms, and to filter out pedestrian-irrelevant style information that compromises identity discriminability and modality invariance of features.

\subsection{Ablation Study}

We perform ablation experiments on the SYSU-MM01 dataset to assess the plausibility and efficacy of each component within the proposed DSFAD. All experiments are conducted under the all-search and single-shot testing mode. The results are presented in Table \ref{Tab:4}, where 'Baseline' represents the base model trained with identity loss and modality-shared enhancement loss.

\begin{table}[!ht]\small
\centering {\caption{Results of Ablation Experiment}\label{Tab:4}
\renewcommand\arraystretch{1.1}
\begin{tabular}{c|cc}
\hline

  & R1 & mAP \\
 \hline

  Baseline & 71.9 & 67.6 \\

  Baseline+DSFA & 77.0 ($\uparrow$5.1) & 73.5($\uparrow$4.9) \\

  Baseline+DSFA+SMFD & 77.9($\uparrow$0.9) & 74.1($\uparrow$0.6) \\

  Baseline+DSFA+SMFD+SCFR & \textbf{78.4}($\uparrow$0.5) & \textbf{74.7}($\uparrow$0.6) \\

  \hline
\end{tabular}}
\end{table}

\textbf{Effectiveness of DSFA.} We propose DSFA to drive the VI-ReID model to extract visible and infrared visual features enriched with pedestrian semantics, and to align them into the common textual embedding space to alleviate the modality discrepancy. As shown in Table \ref{Tab:4}, the incorporation of this module increases Rank-1 accuracy from 71.9\% to 77.0\% and mAP from 67.6\% to 73.5\%, demonstrating its effectiveness in improving cross-modality identity matching.

\textbf{Effectiveness of SMFD.} The proposed SMFD aims to eliminate pedestrian-irrelevant style information such as illumination and color contrast, thereby enhancing the identity discriminability and modality invariance of heterogeneous visual features. Table \ref{Tab:4} demonstrates its effectiveness, with improvements of 0.9\% in Rank-1 accuracy and 0.6\% in mAP, respectively.

\textbf{Effectiveness of SCFR.} SCFR is designed to prevent the loss of pedestrian semantics caused by Instance Normalization (IN) during feature disentanglement. As presented in Table \ref{Tab:4}, integrating SCFR with SMFD further enhances recognition performance, achieving an optimal Rank-1 accuracy of 78.4\% and a mAP of 74.7\%. These results validate the rationale behind SCFR and its effectiveness.

\subsection{Parameters Analysis}

\begin{figure*}[t!]
	\centering
	\includegraphics[width=6.8in,height=1.4in]{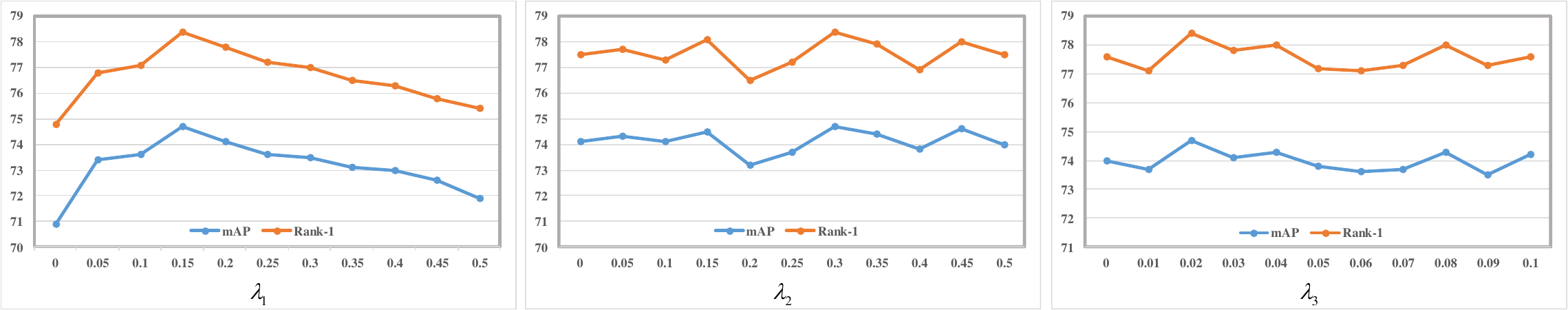}
	\caption{Parameters analysis}
	\label{Fig:4}
\end{figure*}

In the proposed DSFAD framework, in addition to the fundamental identity loss and modality-shared enhancement loss assigned default weights of 1, there are three additional losses: contrastive loss, semantic margin loss, and semantic consistency loss, which are weighted by the hyperparameters $\lambda_{1}$, $\lambda_{2}$, and $\lambda_{3}$, respectively, to balance their relative importance to the overall training loss. We analyze them, with the results presented in Figure \ref{Fig:4}. First, both of them are constrained to be less than 1, as higher values would weaken the effects of identity loss and modality-shared enhancement loss, thereby reducing feature discriminability. Second, $\lambda_{3}$ is significantly smaller than $\lambda_{2}$ because the introduction of Instance Normalization (IN) and the Squeeze-and-Excitation Networks (SE) inevitably modify feature representations. Enforcing excessive semantic consistency between the restored and pre-IN features would interfere with the filtering and recovery functions of IN and SE. Additionally, setting $\lambda_{2}$ and $\lambda_{3}$ to 0 degrades performance, confirming the effectiveness of the developed novel semantic margin and semantic consistency losses. Finally, as shown in Figure \ref{Fig:4}, the optimal values for $\lambda_{1}$, $\lambda_{2}$, and $\lambda_{3}$ are 0.15, 0.3, and 0.02, respectively.

\subsection{Further Discussion}

\textbf{The superiority of diverse textual descriptions.} As discussed in the introduction, a key motivation of this paper is that learnable textual prompts fail to accurately and comprehensively capture pedestrian semantics, while directly employing LLVM-generated textual descriptions results in the model overfitting to fixed semantic patterns. To address these limitations, we propose using description templates with diverse sentence structures to better acquire pedestrian semantics. A qualitative analysis of this approach, presented in Table \ref{Tab:5}, shows that learnable textual prompts achieve a Rank-1 accuracy of 73.6\% and a mAP of 69.8\%. Explicit textual descriptions further enhance Rank-1 accuracy by 1.7\% and mAP by 1.9\%. In comparison, our diverse textual descriptions improve Rank-1 by 3.4\% and mAP by 3.7\%, nearly doubling the gains provided by explicit textual descriptions. These findings validate our motivation and demonstrate the effectiveness of the developed technique.

\begin{table}[!ht]\small
\centering {\caption{Impact of text on performance}\label{Tab:5}
\renewcommand\arraystretch{1.1}
\begin{tabular}{c|cc}
\hline

  & R1 & mAP \\
 \hline

  Textual Prompts & 73.6 & 69.8 \\

  Textual Description & 75.3 ($\uparrow$1.7) & 71.7 ($\uparrow$1.9) \\

  Diverse Textual Descriptions & 77.0 ($\uparrow$3.4) & 73.5 ($\uparrow$3.7) \\

  \hline
\end{tabular}}
\end{table}

\textbf{Model Complexity.} We analyze the model complexity of the proposed DSFAD framework, with the results presented in Table \ref{Tab:6}. First, the baseline model, which solely employs the CLIP pre-trained visual encoder, contains 39.3M parameters and has an inference time of 58s. When DSFA is incorporated, the parameter count increases during training due to the addition of a textual encoder, which has approximately 63.7M parameters. However, since the textual encoder is not required during inference, the model’s parameter count and inference time remain unchanged at test time. Furthermore, when SMFD is introduced to disentangle identity-irrelevant style information, an additional Block4 is incorporated, increasing 15M parameters during training. Nevertheless, as only identity-related pedestrian features are encoded at test time, neither the parameter count nor the inference time is affected during testing. Lastly, integrating the SCFR module introduces an SE network with approximately 0.1M parameters, slightly increasing the model size for both training and inference, while inference time remains unchanged. Overall, the parameter count of our model remains within an acceptable range, enabling deployment on one standard 32GB GPU. Moreover, since the required parameters and inference time remain nearly identical to those of the baseline model during testing, our approach is practical for real-world applications.

\begin{table}[!ht]\small
\centering {\caption{Analysis of model complexity.}\label{Tab:6}
\renewcommand\arraystretch{1.1}
\begin{tabular}{c|c|cc}
\hline
  & Training & \multicolumn{2}{c}{Testing} \\

  & Params & Params & IT\\
 \hline

  Baseline & 39.3M & 39.3M & 58s \\

  Baseline+DSFA & 102.0M & 39.3M & 58s\\

  Baseline+DSFA+SMFD & 117.0M & 39.3M & 58s\\

  Baseline+DSFA+SMFD+SCFR & 117.1M & 39.4M & 58s\\

  \hline
\end{tabular}}
\end{table}

\textbf{Limitations.} In this study, we generate diverse textual descriptions to facilitate the alignment of visible and infrared visual features. However, the random selection of sentence templates introduces instability into the model. A more effective approach may involve adaptively selecting the optimal textual description for each image to guide visual feature learning. Moreover, while LLaVA can readily generate pedestrian descriptions, its introduction still limits the model's practicality. Additionally, the generated descriptions may contain noisy information, particularly for infrared images. A potential solution is to construct a concept set of pedestrian-related visual clues, such as gender, age, hairstyle, and accessories, and leverage them to adaptively select the most relevant attributes, enabling a comprehensive semantic representation of pedestrians without relying on LLaVA. These challenges and opportunities motivate us to conduct more in-depth research in the future.

\section{Conclusion}
In this paper, we proposed an effective Body Shape-aware Textual Alignment (BSaTa) framework to address the modality gap and semantic limitations in VIReID. Unlike prior works that overlook the importance of body shape information, our approach explicitly models and leverages body shape cues through structured textual representations. The proposed BSTA module effectively extracts body shape semantics using human parsing and CLIP-based encoding, while the TVCR ensures alignment between textual and visual body shape features. Furthermore, the integration of Shape-aware Representation Learning (SRL), incorporating multi-text supervision and distribution consistency constraint, enables the visual encoder to learn more discriminative and modality-invariant identity representations. Extensive experiments on the SYSU-MM01 and RegDB datasets demonstrate that our method achieves superior performance, validating the effectiveness of body shape modeling and our BSaTa. In the future, we plan to explore more fine-grained and dynamic textual guidance to further enhance VIReID.


\clearpage
\bibliographystyle{ACM-Reference-Format}
\bibliography{sample-base}


\begin{thebibliography}{66}


\ifx \showCODEN    \undefined \def \showCODEN     #1{\unskip}     \fi
\ifx \showDOI      \undefined \def \showDOI       #1{#1}\fi
\ifx \showISBNx    \undefined \def \showISBNx     #1{\unskip}     \fi
\ifx \showISBNxiii \undefined \def \showISBNxiii  #1{\unskip}     \fi
\ifx \showISSN     \undefined \def \showISSN      #1{\unskip}     \fi
\ifx \showLCCN     \undefined \def \showLCCN      #1{\unskip}     \fi
\ifx \shownote     \undefined \def \shownote      #1{#1}          \fi
\ifx \showarticletitle \undefined \def \showarticletitle #1{#1}   \fi
\ifx \showURL      \undefined \def \showURL       {\relax}        \fi
\providecommand\bibfield[2]{#2}
\providecommand\bibinfo[2]{#2}
\providecommand\natexlab[1]{#1}
\providecommand\showeprint[2][]{arXiv:#2}

\bibitem[Alehdaghi et~al\mbox{.}(2025)]%
        {agpi2}
\bibfield{author}{\bibinfo{person}{Mahdi Alehdaghi}, \bibinfo{person}{Arthur Josi}, \bibinfo{person}{Rafael M.~O. Cruz}, \bibinfo{person}{Pourya Shamsolameli}, {and} \bibinfo{person}{Eric Granger}.} \bibinfo{year}{2025}\natexlab{}.
\newblock \showarticletitle{Adaptive Generation of Privileged Intermediate Information for Visible-Infrared Person Re-Identification}.
\newblock \bibinfo{journal}{\emph{IEEE Transactions on Information Forensics and Security}} (\bibinfo{year}{2025}), \bibinfo{pages}{1--1}.
\newblock
\urldef\tempurl%
\url{https://doi.org/10.1109/TIFS.2025.3541969}
\showDOI{\tempurl}


\bibitem[Chen et~al\mbox{.}(2021)]%
        {nfs}
\bibfield{author}{\bibinfo{person}{Yehansen Chen}, \bibinfo{person}{Lin Wan}, \bibinfo{person}{Zhihang Li}, \bibinfo{person}{Qianyan Jing}, {and} \bibinfo{person}{Zongyuan Sun}.} \bibinfo{year}{2021}\natexlab{}.
\newblock \showarticletitle{Neural feature search for rgb-infrared person re-identification}. In \bibinfo{booktitle}{\emph{Proceedings of the IEEE/CVF conference on computer vision and pattern recognition}}. \bibinfo{pages}{587--597}.
\newblock


\bibitem[Cheng et~al\mbox{.}(2023)]%
        {mbce}
\bibfield{author}{\bibinfo{person}{De Cheng}, \bibinfo{person}{Xiaolong Wang}, \bibinfo{person}{Nannan Wang}, \bibinfo{person}{Zhen Wang}, \bibinfo{person}{Xiaoyu Wang}, {and} \bibinfo{person}{Xinbo Gao}.} \bibinfo{year}{2023}\natexlab{}.
\newblock \showarticletitle{Cross-modality person re-identification with memory-based contrastive embedding}. In \bibinfo{booktitle}{\emph{Proceedings of the AAAI conference on artificial intelligence}}, Vol.~\bibinfo{volume}{37}. \bibinfo{pages}{425--432}.
\newblock


\bibitem[Choi et~al\mbox{.}(2020)]%
        {hicmd}
\bibfield{author}{\bibinfo{person}{Seokeon Choi}, \bibinfo{person}{Sumin Lee}, \bibinfo{person}{Youngeun Kim}, \bibinfo{person}{Taekyung Kim}, {and} \bibinfo{person}{Changick Kim}.} \bibinfo{year}{2020}\natexlab{}.
\newblock \showarticletitle{Hi-CMD: Hierarchical cross-modality disentanglement for visible-infrared person re-identification}. In \bibinfo{booktitle}{\emph{Proceedings of the IEEE/CVF conference on computer vision and pattern recognition}}. \bibinfo{pages}{10257--10266}.
\newblock


\bibitem[Dong et~al\mbox{.}(2024a)]%
        {mvip}
\bibfield{author}{\bibinfo{person}{Neng Dong}, \bibinfo{person}{Shuanglin Yan}, \bibinfo{person}{Hao Tang}, \bibinfo{person}{Jinhui Tang}, {and} \bibinfo{person}{Liyan Zhang}.} \bibinfo{year}{2024}\natexlab{a}.
\newblock \showarticletitle{Multi-view Information Integration and Propagation for occluded person re-identification}.
\newblock \bibinfo{journal}{\emph{Information Fusion}}  \bibinfo{volume}{104} (\bibinfo{year}{2024}), \bibinfo{pages}{102201}.
\newblock


\bibitem[Dong et~al\mbox{.}(2024b)]%
        {eees}
\bibfield{author}{\bibinfo{person}{Neng Dong}, \bibinfo{person}{Shuanglin Yan}, \bibinfo{person}{Liyan Zhang}, {and} \bibinfo{person}{Jinhui Tang}.} \bibinfo{year}{2024}\natexlab{b}.
\newblock \showarticletitle{Embedding and Enriching Explicit Semantics for Visible-Infrared Person Re-Identification}.
\newblock \bibinfo{journal}{\emph{arXiv preprint arXiv:2412.08406}} (\bibinfo{year}{2024}).
\newblock


\bibitem[Dong et~al\mbox{.}(2024c)]%
        {entdnet}
\bibfield{author}{\bibinfo{person}{Neng Dong}, \bibinfo{person}{Liyan Zhang}, \bibinfo{person}{Shuanglin Yan}, \bibinfo{person}{Hao Tang}, {and} \bibinfo{person}{Jinhui Tang}.} \bibinfo{year}{2024}\natexlab{c}.
\newblock \showarticletitle{Erasing, Transforming, and Noising Defense Network for Occluded Person Re-Identification}.
\newblock \bibinfo{journal}{\emph{IEEE Transactions on Circuits and Systems for Video Technology}} \bibinfo{volume}{34}, \bibinfo{number}{6} (\bibinfo{year}{2024}), \bibinfo{pages}{4458--4472}.
\newblock
\urldef\tempurl%
\url{https://doi.org/10.1109/TCSVT.2023.3339167}
\showDOI{\tempurl}


\bibitem[Du et~al\mbox{.}(2022)]%
        {detpro}
\bibfield{author}{\bibinfo{person}{Yu Du}, \bibinfo{person}{Fangyun Wei}, \bibinfo{person}{Zihe Zhang}, \bibinfo{person}{Miaojing Shi}, \bibinfo{person}{Yue Gao}, {and} \bibinfo{person}{Guoqi Li}.} \bibinfo{year}{2022}\natexlab{}.
\newblock \showarticletitle{Learning to prompt for open-vocabulary object detection with vision-language model}. In \bibinfo{booktitle}{\emph{Proceedings of the IEEE/CVF Conference on Computer Vision and Pattern Recognition}}. \bibinfo{pages}{14084--14093}.
\newblock


\bibitem[Feng et~al\mbox{.}(2023)]%
        {sefl}
\bibfield{author}{\bibinfo{person}{Jiawei Feng}, \bibinfo{person}{Ancong Wu}, {and} \bibinfo{person}{Wei-Shi Zheng}.} \bibinfo{year}{2023}\natexlab{}.
\newblock \showarticletitle{Shape-erased feature learning for visible-infrared person re-identification}. In \bibinfo{booktitle}{\emph{Proceedings of the IEEE/CVF conference on computer vision and pattern recognition}}. \bibinfo{pages}{22752--22761}.
\newblock


\bibitem[Gong et~al\mbox{.}(2022)]%
        {cajd}
\bibfield{author}{\bibinfo{person}{Yunpeng Gong}, \bibinfo{person}{Liqing Huang}, {and} \bibinfo{person}{Lifei Chen}.} \bibinfo{year}{2022}\natexlab{}.
\newblock \showarticletitle{Person re-identification method based on color attack and joint defence}. In \bibinfo{booktitle}{\emph{Proceedings of the IEEE/CVF conference on computer vision and pattern recognition}}. \bibinfo{pages}{4313--4322}.
\newblock


\bibitem[Goodfellow et~al\mbox{.}(2020)]%
        {gan}
\bibfield{author}{\bibinfo{person}{Ian Goodfellow}, \bibinfo{person}{Jean Pouget-Abadie}, \bibinfo{person}{Mehdi Mirza}, \bibinfo{person}{Bing Xu}, \bibinfo{person}{David Warde-Farley}, \bibinfo{person}{Sherjil Ozair}, \bibinfo{person}{Aaron Courville}, {and} \bibinfo{person}{Yoshua Bengio}.} \bibinfo{year}{2020}\natexlab{}.
\newblock \showarticletitle{Generative adversarial networks}.
\newblock \bibinfo{journal}{\emph{Commun. ACM}} \bibinfo{volume}{63}, \bibinfo{number}{11} (\bibinfo{year}{2020}), \bibinfo{pages}{139--144}.
\newblock


\bibitem[He et~al\mbox{.}(2016)]%
        {resnet}
\bibfield{author}{\bibinfo{person}{Kaiming He}, \bibinfo{person}{Xiangyu Zhang}, \bibinfo{person}{Shaoqing Ren}, {and} \bibinfo{person}{Jian Sun}.} \bibinfo{year}{2016}\natexlab{}.
\newblock \showarticletitle{Deep residual learning for image recognition}. In \bibinfo{booktitle}{\emph{Proceedings of the IEEE conference on computer vision and pattern recognition}}. \bibinfo{pages}{770--778}.
\newblock


\bibitem[Hu et~al\mbox{.}(2018)]%
        {se}
\bibfield{author}{\bibinfo{person}{Jie Hu}, \bibinfo{person}{Li Shen}, {and} \bibinfo{person}{Gang Sun}.} \bibinfo{year}{2018}\natexlab{}.
\newblock \showarticletitle{Squeeze-and-excitation networks}. In \bibinfo{booktitle}{\emph{Proceedings of the IEEE conference on computer vision and pattern recognition}}. \bibinfo{pages}{7132--7141}.
\newblock


\bibitem[Huang and Belongie(2017)]%
        {in}
\bibfield{author}{\bibinfo{person}{Xun Huang} {and} \bibinfo{person}{Serge Belongie}.} \bibinfo{year}{2017}\natexlab{}.
\newblock \showarticletitle{Arbitrary style transfer in real-time with adaptive instance normalization}. In \bibinfo{booktitle}{\emph{Proceedings of the IEEE international conference on computer vision}}. \bibinfo{pages}{1501--1510}.
\newblock


\bibitem[Jaderberg et~al\mbox{.}(2015)]%
        {transformer}
\bibfield{author}{\bibinfo{person}{Max Jaderberg}, \bibinfo{person}{Karen Simonyan}, \bibinfo{person}{Andrew Zisserman}, {et~al\mbox{.}}} \bibinfo{year}{2015}\natexlab{}.
\newblock \showarticletitle{Spatial transformer networks}.
\newblock \bibinfo{journal}{\emph{Advances in neural information processing systems}}  \bibinfo{volume}{28} (\bibinfo{year}{2015}).
\newblock


\bibitem[Jeong et~al\mbox{.}(2023)]%
        {winclip}
\bibfield{author}{\bibinfo{person}{Jongheon Jeong}, \bibinfo{person}{Yang Zou}, \bibinfo{person}{Taewan Kim}, \bibinfo{person}{Dongqing Zhang}, \bibinfo{person}{Avinash Ravichandran}, {and} \bibinfo{person}{Onkar Dabeer}.} \bibinfo{year}{2023}\natexlab{}.
\newblock \showarticletitle{Winclip: Zero-/few-shot anomaly classification and segmentation}. In \bibinfo{booktitle}{\emph{Proceedings of the IEEE/CVF Conference on Computer Vision and Pattern Recognition}}. \bibinfo{pages}{19606--19616}.
\newblock


\bibitem[Jiang et~al\mbox{.}(2022)]%
        {cmt}
\bibfield{author}{\bibinfo{person}{Kongzhu Jiang}, \bibinfo{person}{Tianzhu Zhang}, \bibinfo{person}{Xiang Liu}, \bibinfo{person}{Bingqiao Qian}, \bibinfo{person}{Yongdong Zhang}, {and} \bibinfo{person}{Feng Wu}.} \bibinfo{year}{2022}\natexlab{}.
\newblock \showarticletitle{Cross-modality transformer for visible-infrared person re-identification}. In \bibinfo{booktitle}{\emph{European Conference on Computer Vision}}. Springer, \bibinfo{pages}{480--496}.
\newblock


\bibitem[Jin et~al\mbox{.}(2020)]%
        {snr}
\bibfield{author}{\bibinfo{person}{Xin Jin}, \bibinfo{person}{Cuiling Lan}, \bibinfo{person}{Wenjun Zeng}, \bibinfo{person}{Zhibo Chen}, {and} \bibinfo{person}{Li Zhang}.} \bibinfo{year}{2020}\natexlab{}.
\newblock \showarticletitle{Style normalization and restitution for generalizable person re-identification}. In \bibinfo{booktitle}{\emph{Proceedings of the IEEE/CVF conference on computer vision and pattern recognition}}. \bibinfo{pages}{3143--3152}.
\newblock


\bibitem[Kingma(2014)]%
        {adam}
\bibfield{author}{\bibinfo{person}{Diederik~P Kingma}.} \bibinfo{year}{2014}\natexlab{}.
\newblock \showarticletitle{Adam: A method for stochastic optimization}.
\newblock \bibinfo{journal}{\emph{arXiv preprint arXiv:1412.6980}} (\bibinfo{year}{2014}).
\newblock


\bibitem[Li et~al\mbox{.}(2020)]%
        {xiv-reid}
\bibfield{author}{\bibinfo{person}{Diangang Li}, \bibinfo{person}{Xing Wei}, \bibinfo{person}{Xiaopeng Hong}, {and} \bibinfo{person}{Yihong Gong}.} \bibinfo{year}{2020}\natexlab{}.
\newblock \showarticletitle{Infrared-visible cross-modal person re-identification with an x modality}. In \bibinfo{booktitle}{\emph{Proceedings of the AAAI conference on artificial intelligence}}, Vol.~\bibinfo{volume}{34}. \bibinfo{pages}{4610--4617}.
\newblock


\bibitem[Li et~al\mbox{.}(2021)]%
        {tal}
\bibfield{author}{\bibinfo{person}{Huafeng Li}, \bibinfo{person}{Neng Dong}, \bibinfo{person}{Zhengtao Yu}, \bibinfo{person}{Dapeng Tao}, {and} \bibinfo{person}{Guanqiu Qi}.} \bibinfo{year}{2021}\natexlab{}.
\newblock \showarticletitle{Triple adversarial learning and multi-view imaginative reasoning for unsupervised domain adaptation person re-identification}.
\newblock \bibinfo{journal}{\emph{IEEE Transactions on Circuits and Systems for Video Technology}} \bibinfo{volume}{32}, \bibinfo{number}{5} (\bibinfo{year}{2021}), \bibinfo{pages}{2814--2830}.
\newblock


\bibitem[Li et~al\mbox{.}(2023)]%
        {clipreid}
\bibfield{author}{\bibinfo{person}{Siyuan Li}, \bibinfo{person}{Li Sun}, {and} \bibinfo{person}{Qingli Li}.} \bibinfo{year}{2023}\natexlab{}.
\newblock \showarticletitle{CLIP-ReID: exploiting vision-language model for image re-identification without concrete text labels}. In \bibinfo{booktitle}{\emph{Proceedings of the AAAI Conference on Artificial Intelligence}}, Vol.~\bibinfo{volume}{37}. \bibinfo{pages}{1405--1413}.
\newblock


\bibitem[Lin et~al\mbox{.}(2022)]%
        {lin2022learning}
\bibfield{author}{\bibinfo{person}{Xinyu Lin}, \bibinfo{person}{Jinxing Li}, \bibinfo{person}{Zeyu Ma}, \bibinfo{person}{Huafeng Li}, \bibinfo{person}{Shuang Li}, \bibinfo{person}{Kaixiong Xu}, \bibinfo{person}{Guangming Lu}, {and} \bibinfo{person}{David Zhang}.} \bibinfo{year}{2022}\natexlab{}.
\newblock \showarticletitle{Learning modal-invariant and temporal-memory for video-based visible-infrared person re-identification}. In \bibinfo{booktitle}{\emph{IEEE Conference on Computer Vision and Pattern Recognition (CVPR)}}. \bibinfo{pages}{20973--20982}.
\newblock


\bibitem[Liu et~al\mbox{.}(2023)]%
        {llava}
\bibfield{author}{\bibinfo{person}{Haotian Liu}, \bibinfo{person}{Chunyuan Li}, \bibinfo{person}{Qingyang Wu}, {and} \bibinfo{person}{Yong~Jae Lee}.} \bibinfo{year}{2023}\natexlab{}.
\newblock \showarticletitle{Visual Instruction Tuning}. In \bibinfo{booktitle}{\emph{Advances in Neural Information Processing Systems}}, \bibfield{editor}{\bibinfo{person}{A.~Oh}, \bibinfo{person}{T.~Naumann}, \bibinfo{person}{A.~Globerson}, \bibinfo{person}{K.~Saenko}, \bibinfo{person}{M.~Hardt}, {and} \bibinfo{person}{S.~Levine}} (Eds.), Vol.~\bibinfo{volume}{36}. \bibinfo{publisher}{Curran Associates, Inc.}, \bibinfo{pages}{34892--34916}.
\newblock


\bibitem[Liu et~al\mbox{.}(2022)]%
        {maum}
\bibfield{author}{\bibinfo{person}{Jialun Liu}, \bibinfo{person}{Yifan Sun}, \bibinfo{person}{Feng Zhu}, \bibinfo{person}{Hongbin Pei}, \bibinfo{person}{Yi Yang}, {and} \bibinfo{person}{Wenhui Li}.} \bibinfo{year}{2022}\natexlab{}.
\newblock \showarticletitle{Learning memory-augmented unidirectional metrics for cross-modality person re-identification}. In \bibinfo{booktitle}{\emph{Proceedings of the IEEE/CVF conference on computer vision and pattern recognition}}. \bibinfo{pages}{19366--19375}.
\newblock


\bibitem[Liu et~al\mbox{.}(2025)]%
        {cscl}
\bibfield{author}{\bibinfo{person}{Min Liu}, \bibinfo{person}{Zhu Zhang}, \bibinfo{person}{Yuan Bian}, \bibinfo{person}{Xueping Wang}, \bibinfo{person}{Yeqing Sun}, \bibinfo{person}{Baida Zhang}, {and} \bibinfo{person}{Yaonan Wang}.} \bibinfo{year}{2025}\natexlab{}.
\newblock \showarticletitle{Cross-Modality Semantic Consistency Learning for Visible-Infrared Person Re-Identification}.
\newblock \bibinfo{journal}{\emph{IEEE Transactions on Multimedia}}  \bibinfo{volume}{27} (\bibinfo{year}{2025}), \bibinfo{pages}{568--580}.
\newblock
\urldef\tempurl%
\url{https://doi.org/10.1109/TMM.2024.3521843}
\showDOI{\tempurl}


\bibitem[Lu et~al\mbox{.}(2023)]%
        {mse}
\bibfield{author}{\bibinfo{person}{Hu Lu}, \bibinfo{person}{Xuezhang Zou}, {and} \bibinfo{person}{Pingping Zhang}.} \bibinfo{year}{2023}\natexlab{}.
\newblock \showarticletitle{Learning progressive modality-shared transformers for effective visible-infrared person re-identification}. In \bibinfo{booktitle}{\emph{Proceedings of the AAAI conference on artificial intelligence}}, Vol.~\bibinfo{volume}{37}. \bibinfo{pages}{1835--1843}.
\newblock


\bibitem[Lu et~al\mbox{.}(2025)]%
        {dpmf}
\bibfield{author}{\bibinfo{person}{Zefeng Lu}, \bibinfo{person}{Ronghao Lin}, {and} \bibinfo{person}{Haifeng Hu}.} \bibinfo{year}{2025}\natexlab{}.
\newblock \showarticletitle{Disentangling Modality and Posture Factors: Memory-Attention and Orthogonal Decomposition for Visible-Infrared Person Re-Identification}.
\newblock \bibinfo{journal}{\emph{IEEE Transactions on Neural Networks and Learning Systems}} \bibinfo{volume}{36}, \bibinfo{number}{3} (\bibinfo{year}{2025}), \bibinfo{pages}{5494--5508}.
\newblock
\urldef\tempurl%
\url{https://doi.org/10.1109/TNNLS.2024.3384023}
\showDOI{\tempurl}


\bibitem[Luo et~al\mbox{.}(2019)]%
        {bagtricks}
\bibfield{author}{\bibinfo{person}{Hao Luo}, \bibinfo{person}{Youzhi Gu}, \bibinfo{person}{Xingyu Liao}, \bibinfo{person}{Shenqi Lai}, {and} \bibinfo{person}{Wei Jiang}.} \bibinfo{year}{2019}\natexlab{}.
\newblock \showarticletitle{Bag of Tricks and a Strong Baseline for Deep Person Re-Identification}. In \bibinfo{booktitle}{\emph{Proceedings of the IEEE Conference on Computer Vision and Pattern Recognition Workshops (CVPRW)}}. \bibinfo{pages}{1487--1495}.
\newblock
\urldef\tempurl%
\url{https://doi.org/10.1109/CVPRW.2019.00190}
\showDOI{\tempurl}


\bibitem[Nguyen et~al\mbox{.}(2017)]%
        {regdb}
\bibfield{author}{\bibinfo{person}{Dat~Tien Nguyen}, \bibinfo{person}{Hyung~Gil Hong}, \bibinfo{person}{Ki~Wan Kim}, {and} \bibinfo{person}{Kang~Ryoung Park}.} \bibinfo{year}{2017}\natexlab{}.
\newblock \showarticletitle{Person recognition system based on a combination of body images from visible light and thermal cameras}.
\newblock \bibinfo{journal}{\emph{Sensors}} \bibinfo{volume}{17}, \bibinfo{number}{3} (\bibinfo{year}{2017}), \bibinfo{pages}{605}.
\newblock


\bibitem[OpenAI(2022)]%
        {chatgpt}
\bibfield{author}{\bibinfo{person}{OpenAI}.} \bibinfo{year}{2022}\natexlab{}.
\newblock \showarticletitle{Chatgpt}. In \bibinfo{booktitle}{\emph{https://openai.com/blog/chatgpt/}}.
\newblock


\bibitem[Pan et~al\mbox{.}(2024)]%
        {acd}
\bibfield{author}{\bibinfo{person}{Honghu Pan}, \bibinfo{person}{Wenjie Pei}, \bibinfo{person}{Xin Li}, {and} \bibinfo{person}{Zhenyu He}.} \bibinfo{year}{2024}\natexlab{}.
\newblock \showarticletitle{Unified Conditional Image Generation for Visible-Infrared Person Re-Identification}.
\newblock \bibinfo{journal}{\emph{IEEE Transactions on Information Forensics and Security}} (\bibinfo{year}{2024}).
\newblock


\bibitem[Qi et~al\mbox{.}(2025)]%
        {argn}
\bibfield{author}{\bibinfo{person}{Mengzan Qi}, \bibinfo{person}{Sixian Chan}, \bibinfo{person}{Chen Hang}, \bibinfo{person}{Guixu Zhang}, \bibinfo{person}{Tieyong Zeng}, {and} \bibinfo{person}{Zhi Li}.} \bibinfo{year}{2025}\natexlab{}.
\newblock \showarticletitle{Auxiliary Representation Guided Network for Visible-Infrared Person Re-Identification}.
\newblock \bibinfo{journal}{\emph{IEEE Transactions on Multimedia}}  \bibinfo{volume}{27} (\bibinfo{year}{2025}), \bibinfo{pages}{340--355}.
\newblock
\urldef\tempurl%
\url{https://doi.org/10.1109/TMM.2024.3521773}
\showDOI{\tempurl}


\bibitem[Qiu et~al\mbox{.}(2024)]%
        {hosnet}
\bibfield{author}{\bibinfo{person}{Liuxiang Qiu}, \bibinfo{person}{Si Chen}, \bibinfo{person}{Yan Yan}, \bibinfo{person}{Jing-Hao Xue}, \bibinfo{person}{Da-Han Wang}, {and} \bibinfo{person}{Shunzhi Zhu}.} \bibinfo{year}{2024}\natexlab{}.
\newblock \showarticletitle{High-order structure based middle-feature learning for visible-infrared person re-identification}. In \bibinfo{booktitle}{\emph{Proceedings of the AAAI Conference on Artificial Intelligence}}, Vol.~\bibinfo{volume}{38}. \bibinfo{pages}{4596--4604}.
\newblock


\bibitem[Radford et~al\mbox{.}(2021)]%
        {clip}
\bibfield{author}{\bibinfo{person}{Alec Radford}, \bibinfo{person}{Jong~Wook Kim}, \bibinfo{person}{Chris Hallacy}, \bibinfo{person}{Aditya Ramesh}, \bibinfo{person}{Gabriel Goh}, \bibinfo{person}{Sandhini Agarwal}, \bibinfo{person}{Girish Sastry}, \bibinfo{person}{Amanda Askell}, \bibinfo{person}{Pamela Mishkin}, \bibinfo{person}{Jack Clark}, {et~al\mbox{.}}} \bibinfo{year}{2021}\natexlab{}.
\newblock \showarticletitle{Learning transferable visual models from natural language supervision}. In \bibinfo{booktitle}{\emph{International conference on machine learning}}. PMLR, \bibinfo{pages}{8748--8763}.
\newblock


\bibitem[Shen et~al\mbox{.}(2025)]%
        {shen2025imagdressing}
\bibfield{author}{\bibinfo{person}{Fei Shen}, \bibinfo{person}{Xin Jiang}, \bibinfo{person}{Xin He}, \bibinfo{person}{Hu Ye}, \bibinfo{person}{Cong Wang}, \bibinfo{person}{Xiaoyu Du}, \bibinfo{person}{Zechao Li}, {and} \bibinfo{person}{Jinhui Tang}.} \bibinfo{year}{2025}\natexlab{}.
\newblock \showarticletitle{Imagdressing-v1: Customizable virtual dressing}. In \bibinfo{booktitle}{\emph{Proceedings of the AAAI Conference on Artificial Intelligence}}, Vol.~\bibinfo{volume}{39}. \bibinfo{pages}{6795--6804}.
\newblock


\bibitem[Shen et~al\mbox{.}(2023)]%
        {shen2023pbsl}
\bibfield{author}{\bibinfo{person}{Fei Shen}, \bibinfo{person}{Xiangbo Shu}, \bibinfo{person}{Xiaoyu Du}, {and} \bibinfo{person}{Jinhui Tang}.} \bibinfo{year}{2023}\natexlab{}.
\newblock \showarticletitle{Pedestrian-specific Bipartite-aware Similarity Learning for Text-based Person Retrieval}. In \bibinfo{booktitle}{\emph{Proceedings of the 31th ACM International Conference on Multimedia}}.
\newblock


\bibitem[Shen and Tang(2024)]%
        {shen2024imagpose}
\bibfield{author}{\bibinfo{person}{Fei Shen} {and} \bibinfo{person}{Jinhui Tang}.} \bibinfo{year}{2024}\natexlab{}.
\newblock \showarticletitle{Imagpose: A unified conditional framework for pose-guided person generation}.
\newblock \bibinfo{journal}{\emph{Advances in neural information processing systems}}  \bibinfo{volume}{37} (\bibinfo{year}{2024}), \bibinfo{pages}{6246--6266}.
\newblock


\bibitem[Shi et~al\mbox{.}(2023)]%
        {dpli}
\bibfield{author}{\bibinfo{person}{Jiangming Shi}, \bibinfo{person}{Yachao Zhang}, \bibinfo{person}{Xiangbo Yin}, \bibinfo{person}{Yuan Xie}, \bibinfo{person}{Zhizhong Zhang}, \bibinfo{person}{Jianping Fan}, \bibinfo{person}{Zhongchao Shi}, {and} \bibinfo{person}{Yanyun Qu}.} \bibinfo{year}{2023}\natexlab{}.
\newblock \showarticletitle{Dual pseudo-labels interactive self-training for semi-supervised visible-infrared person re-identification}. In \bibinfo{booktitle}{\emph{Proceedings of the IEEE/CVF International Conference on Computer Vision}}. \bibinfo{pages}{11218--11228}.
\newblock


\bibitem[Sun et~al\mbox{.}(2022)]%
        {dclnet}
\bibfield{author}{\bibinfo{person}{Hanzhe Sun}, \bibinfo{person}{Jun Liu}, \bibinfo{person}{Zhizhong Zhang}, \bibinfo{person}{Chengjie Wang}, \bibinfo{person}{Yanyun Qu}, \bibinfo{person}{Yuan Xie}, {and} \bibinfo{person}{Lizhuang Ma}.} \bibinfo{year}{2022}\natexlab{}.
\newblock \showarticletitle{Not all pixels are matched: Dense contrastive learning for cross-modality person re-identification}. In \bibinfo{booktitle}{\emph{Proceedings of the 30th ACM international conference on multimedia}}. \bibinfo{pages}{5333--5341}.
\newblock


\bibitem[Sun et~al\mbox{.}(2021)]%
        {pcb}
\bibfield{author}{\bibinfo{person}{Yifan Sun}, \bibinfo{person}{Liang Zheng}, \bibinfo{person}{Yali Li}, \bibinfo{person}{Yi Yang}, \bibinfo{person}{Qi Tian}, {and} \bibinfo{person}{Shengjin Wang}.} \bibinfo{year}{2021}\natexlab{}.
\newblock \showarticletitle{Learning Part-based Convolutional Features for Person Re-Identification}.
\newblock \bibinfo{journal}{\emph{IEEE Transactions on Pattern Analysis and Machine Intelligence}} \bibinfo{volume}{43}, \bibinfo{number}{3} (\bibinfo{year}{2021}), \bibinfo{pages}{902--917}.
\newblock
\urldef\tempurl%
\url{https://doi.org/10.1109/TPAMI.2019.2938523}
\showDOI{\tempurl}


\bibitem[Tan et~al\mbox{.}(2024)]%
        {rle}
\bibfield{author}{\bibinfo{person}{Lei Tan}, \bibinfo{person}{Yukang Zhang}, \bibinfo{person}{Keke Han}, \bibinfo{person}{Pingyang Dai}, \bibinfo{person}{Yan Zhang}, \bibinfo{person}{YONGJIAN WU}, {and} \bibinfo{person}{Rongrong Ji}.} \bibinfo{year}{2024}\natexlab{}.
\newblock \showarticletitle{{RLE}: A Unified Perspective of Data Augmentation for Cross-Spectral Re-Identification}. In \bibinfo{booktitle}{\emph{The Thirty-eighth Annual Conference on Neural Information Processing Systems}}.
\newblock
\urldef\tempurl%
\url{https://openreview.net/forum?id=Ok6jSSxzfj}
\showURL{%
\tempurl}


\bibitem[Tang et~al\mbox{.}(2024)]%
        {tang2024divide}
\bibfield{author}{\bibinfo{person}{Hao Tang}, \bibinfo{person}{Zechao Li}, \bibinfo{person}{Dong Zhang}, \bibinfo{person}{Shengfeng He}, {and} \bibinfo{person}{Jinhui Tang}.} \bibinfo{year}{2024}\natexlab{}.
\newblock \showarticletitle{Divide-and-Conquer: Confluent Triple-Flow Network for RGB-T Salient Object Detection}.
\newblock \bibinfo{journal}{\emph{IEEE Transactions on Pattern Analysis and Machine Intelligence}} (\bibinfo{year}{2024}).
\newblock


\bibitem[Tang et~al\mbox{.}(2023)]%
        {m3net}
\bibfield{author}{\bibinfo{person}{Hao Tang}, \bibinfo{person}{Jun Liu}, \bibinfo{person}{Shuanglin Yan}, \bibinfo{person}{Rui Yan}, \bibinfo{person}{Zechao Li}, {and} \bibinfo{person}{Jinhui Tang}.} \bibinfo{year}{2023}\natexlab{}.
\newblock \showarticletitle{M3Net: Multi-view Encoding, Matching, and Fusion for Few-shot Fine-grained Action Recognition}. In \bibinfo{booktitle}{\emph{ACM International Conference on Multimedia (ACM MM)}}. \bibinfo{pages}{1719–1728}.
\newblock


\bibitem[Wang et~al\mbox{.}(2019b)]%
        {aligngan}
\bibfield{author}{\bibinfo{person}{Guan'an Wang}, \bibinfo{person}{Tianzhu Zhang}, \bibinfo{person}{Jian Cheng}, \bibinfo{person}{Si Liu}, \bibinfo{person}{Yang Yang}, {and} \bibinfo{person}{Zengguang Hou}.} \bibinfo{year}{2019}\natexlab{b}.
\newblock \showarticletitle{RGB-infrared cross-modality person re-identification via joint pixel and feature alignment}. In \bibinfo{booktitle}{\emph{Proceedings of the IEEE/CVF International Conference on Computer Vision}}. \bibinfo{pages}{3623--3632}.
\newblock


\bibitem[Wang et~al\mbox{.}(2022b)]%
        {wang2022body}
\bibfield{author}{\bibinfo{person}{Yiming Wang}, \bibinfo{person}{Guanqiu Qi}, \bibinfo{person}{Shuang Li}, \bibinfo{person}{Yi Chai}, {and} \bibinfo{person}{Huafeng Li}.} \bibinfo{year}{2022}\natexlab{b}.
\newblock \showarticletitle{Body part-level domain alignment for domain-adaptive person re-identification with transformer framework}.
\newblock \bibinfo{journal}{\emph{IEEE Transactions on Information Forensics and Security}}  \bibinfo{volume}{17} (\bibinfo{year}{2022}), \bibinfo{pages}{3321--3334}.
\newblock


\bibitem[Wang et~al\mbox{.}(2022a)]%
        {cris}
\bibfield{author}{\bibinfo{person}{Zhaoqing Wang}, \bibinfo{person}{Yu Lu}, \bibinfo{person}{Qiang Li}, \bibinfo{person}{Xunqiang Tao}, \bibinfo{person}{Yandong Guo}, \bibinfo{person}{Mingming Gong}, {and} \bibinfo{person}{Tongliang Liu}.} \bibinfo{year}{2022}\natexlab{a}.
\newblock \showarticletitle{Cris: Clip-driven referring image segmentation}. In \bibinfo{booktitle}{\emph{Proceedings of the IEEE/CVF conference on computer vision and pattern recognition}}. \bibinfo{pages}{11686--11695}.
\newblock


\bibitem[Wang et~al\mbox{.}(2019a)]%
        {d2rl}
\bibfield{author}{\bibinfo{person}{Zhixiang Wang}, \bibinfo{person}{Zheng Wang}, \bibinfo{person}{Yinqiang Zheng}, \bibinfo{person}{Yung-Yu Chuang}, {and} \bibinfo{person}{Shin'ichi Satoh}.} \bibinfo{year}{2019}\natexlab{a}.
\newblock \showarticletitle{Learning to reduce dual-level discrepancy for infrared-visible person re-identification}. In \bibinfo{booktitle}{\emph{Proceedings of the IEEE/CVF conference on computer vision and pattern recognition}}. \bibinfo{pages}{618--626}.
\newblock


\bibitem[Wu et~al\mbox{.}(2017)]%
        {vireid}
\bibfield{author}{\bibinfo{person}{Ancong Wu}, \bibinfo{person}{Wei-Shi Zheng}, \bibinfo{person}{Hong-Xing Yu}, \bibinfo{person}{Shaogang Gong}, {and} \bibinfo{person}{Jianhuang Lai}.} \bibinfo{year}{2017}\natexlab{}.
\newblock \showarticletitle{RGB-Infrared Cross-Modality Person Re-identification}. In \bibinfo{booktitle}{\emph{2017 IEEE International Conference on Computer Vision (ICCV)}}. \bibinfo{pages}{5390--5399}.
\newblock
\urldef\tempurl%
\url{https://doi.org/10.1109/ICCV.2017.575}
\showDOI{\tempurl}


\bibitem[Wu et~al\mbox{.}(2023)]%
        {cal}
\bibfield{author}{\bibinfo{person}{Jianbing Wu}, \bibinfo{person}{Hong Liu}, \bibinfo{person}{Yuxin Su}, \bibinfo{person}{Wei Shi}, {and} \bibinfo{person}{Hao Tang}.} \bibinfo{year}{2023}\natexlab{}.
\newblock \showarticletitle{Learning concordant attention via target-aware alignment for visible-infrared person re-identification}. In \bibinfo{booktitle}{\emph{Proceedings of the IEEE/CVF international conference on computer vision}}. \bibinfo{pages}{11122--11131}.
\newblock


\bibitem[Wu et~al\mbox{.}(2021)]%
        {mpanet}
\bibfield{author}{\bibinfo{person}{Qiong Wu}, \bibinfo{person}{Pingyang Dai}, \bibinfo{person}{Jie Chen}, \bibinfo{person}{Chia-Wen Lin}, \bibinfo{person}{Yongjian Wu}, \bibinfo{person}{Feiyue Huang}, \bibinfo{person}{Bineng Zhong}, {and} \bibinfo{person}{Rongrong Ji}.} \bibinfo{year}{2021}\natexlab{}.
\newblock \showarticletitle{Discover cross-modality nuances for visible-infrared person re-identification}. In \bibinfo{booktitle}{\emph{Proceedings of the IEEE/CVF conference on computer vision and pattern recognition}}. \bibinfo{pages}{4330--4339}.
\newblock


\bibitem[Yan et~al\mbox{.}(2025)]%
        {trimatch}
\bibfield{author}{\bibinfo{person}{Shuanglin Yan}, \bibinfo{person}{Neng Dong}, \bibinfo{person}{Shuang Li}, {and} \bibinfo{person}{Huafeng Li}.} \bibinfo{year}{2025}\natexlab{}.
\newblock \showarticletitle{TriMatch: Triple Matching for Text-to-Image Person Re-Identification}.
\newblock \bibinfo{journal}{\emph{IEEE Signal Processing Letters}}  \bibinfo{volume}{32} (\bibinfo{year}{2025}), \bibinfo{pages}{806--810}.
\newblock


\bibitem[Yan et~al\mbox{.}(2023a)]%
        {LCR2S}
\bibfield{author}{\bibinfo{person}{Shuanglin Yan}, \bibinfo{person}{Neng Dong}, \bibinfo{person}{Jun Liu}, \bibinfo{person}{Liyan Zhang}, {and} \bibinfo{person}{Jinhui Tang}.} \bibinfo{year}{2023}\natexlab{a}.
\newblock \showarticletitle{Learning Comprehensive Representations with Richer Self for Text-to-Image Person Re-Identification}. In \bibinfo{booktitle}{\emph{ACM International Conference on Multimedia (ACM MM)}}. \bibinfo{pages}{6202–6211}.
\newblock


\bibitem[Yan et~al\mbox{.}(2023b)]%
        {cfine}
\bibfield{author}{\bibinfo{person}{Shuanglin Yan}, \bibinfo{person}{Neng Dong}, \bibinfo{person}{Liyan Zhang}, {and} \bibinfo{person}{Jinhui Tang}.} \bibinfo{year}{2023}\natexlab{b}.
\newblock \showarticletitle{CLIP-Driven Fine-Grained Text-Image Person Re-Identification}.
\newblock \bibinfo{journal}{\emph{IEEE Transactions on Image Processing}}  \bibinfo{volume}{32} (\bibinfo{year}{2023}), \bibinfo{pages}{6032--6046}.
\newblock
\urldef\tempurl%
\url{https://doi.org/10.1109/TIP.2023.3327924}
\showDOI{\tempurl}


\bibitem[Yan et~al\mbox{.}(2024)]%
        {propot}
\bibfield{author}{\bibinfo{person}{Shuanglin Yan}, \bibinfo{person}{Jun Liu}, \bibinfo{person}{Neng Dong}, \bibinfo{person}{Liyan Zhang}, {and} \bibinfo{person}{Jinhui Tang}.} \bibinfo{year}{2024}\natexlab{}.
\newblock \showarticletitle{Prototypical Prompting for Text-to-image Person Re-identification}. In \bibinfo{booktitle}{\emph{ACM International Conference on Multimedia (ACM MM)}}. \bibinfo{pages}{2331–2340}.
\newblock


\bibitem[Yan et~al\mbox{.}(2023c)]%
        {manet}
\bibfield{author}{\bibinfo{person}{Shuanglin Yan}, \bibinfo{person}{Hao Tang}, \bibinfo{person}{Liyan Zhang}, {and} \bibinfo{person}{Jinhui Tang}.} \bibinfo{year}{2023}\natexlab{c}.
\newblock \showarticletitle{Image-Specific Information Suppression and Implicit Local Alignment for Text-Based Person Search}.
\newblock \bibinfo{journal}{\emph{IEEE Transactions on Neural Networks and Learning Systems}} (\bibinfo{year}{2023}), \bibinfo{pages}{1--14}.
\newblock
\urldef\tempurl%
\url{https://doi.org/10.1109/TNNLS.2023.3310118}
\showDOI{\tempurl}


\bibitem[Yang et~al\mbox{.}(2022)]%
        {dart}
\bibfield{author}{\bibinfo{person}{Mouxing Yang}, \bibinfo{person}{Zhenyu Huang}, \bibinfo{person}{Peng Hu}, \bibinfo{person}{Taihao Li}, \bibinfo{person}{Jiancheng Lv}, {and} \bibinfo{person}{Xi Peng}.} \bibinfo{year}{2022}\natexlab{}.
\newblock \showarticletitle{Learning with twin noisy labels for visible-infrared person re-identification}. In \bibinfo{booktitle}{\emph{Proceedings of the IEEE/CVF conference on computer vision and pattern recognition}}. \bibinfo{pages}{14308--14317}.
\newblock


\bibitem[Ye et~al\mbox{.}(2023)]%
        {caj+}
\bibfield{author}{\bibinfo{person}{Mang Ye}, \bibinfo{person}{Zesen Wu}, \bibinfo{person}{Cuiqun Chen}, {and} \bibinfo{person}{Bo Du}.} \bibinfo{year}{2023}\natexlab{}.
\newblock \showarticletitle{Channel augmentation for visible-infrared re-identification}.
\newblock \bibinfo{journal}{\emph{IEEE Transactions on Pattern Analysis and Machine Intelligence}} \bibinfo{volume}{46}, \bibinfo{number}{4} (\bibinfo{year}{2023}), \bibinfo{pages}{2299--2315}.
\newblock


\bibitem[Yu et~al\mbox{.}(2023)]%
        {mun}
\bibfield{author}{\bibinfo{person}{Hao Yu}, \bibinfo{person}{Xu Cheng}, \bibinfo{person}{Wei Peng}, \bibinfo{person}{Weihao Liu}, {and} \bibinfo{person}{Guoying Zhao}.} \bibinfo{year}{2023}\natexlab{}.
\newblock \showarticletitle{Modality unifying network for visible-infrared person re-identification}. In \bibinfo{booktitle}{\emph{Proceedings of the IEEE/CVF international conference on computer vision}}. \bibinfo{pages}{11185--11195}.
\newblock


\bibitem[Yu et~al\mbox{.}(2025)]%
        {csdn}
\bibfield{author}{\bibinfo{person}{Xiaoyan Yu}, \bibinfo{person}{Neng Dong}, \bibinfo{person}{Liehuang Zhu}, \bibinfo{person}{Hao Peng}, {and} \bibinfo{person}{Dapeng Tao}.} \bibinfo{year}{2025}\natexlab{}.
\newblock \showarticletitle{CLIP-Driven Semantic Discovery Network for Visible-Infrared Person Re-Identification}.
\newblock \bibinfo{journal}{\emph{IEEE Transactions on Multimedia}} (\bibinfo{year}{2025}), \bibinfo{pages}{1--13}.
\newblock
\urldef\tempurl%
\url{https://doi.org/10.1109/TMM.2025.3535353}
\showDOI{\tempurl}


\bibitem[Zhang et~al\mbox{.}(2023)]%
        {protohpe}
\bibfield{author}{\bibinfo{person}{Guiwei Zhang}, \bibinfo{person}{Yongfei Zhang}, {and} \bibinfo{person}{Zichang Tan}.} \bibinfo{year}{2023}\natexlab{}.
\newblock \showarticletitle{Protohpe: Prototype-guided high-frequency patch enhancement for visible-infrared person re-identification}. In \bibinfo{booktitle}{\emph{Proceedings of the 31st ACM International Conference on Multimedia}}. \bibinfo{pages}{944--954}.
\newblock


\bibitem[Zhang et~al\mbox{.}(2022a)]%
        {fmcnet}
\bibfield{author}{\bibinfo{person}{Qiang Zhang}, \bibinfo{person}{Changzhou Lai}, \bibinfo{person}{Jianan Liu}, \bibinfo{person}{Nianchang Huang}, {and} \bibinfo{person}{Jungong Han}.} \bibinfo{year}{2022}\natexlab{a}.
\newblock \showarticletitle{Fmcnet: Feature-level modality compensation for visible-infrared person re-identification}. In \bibinfo{booktitle}{\emph{Proceedings of the IEEE/CVF conference on computer vision and pattern recognition}}. \bibinfo{pages}{7349--7358}.
\newblock


\bibitem[Zhang and Wang(2023)]%
        {deen}
\bibfield{author}{\bibinfo{person}{Yukang Zhang} {and} \bibinfo{person}{Hanzi Wang}.} \bibinfo{year}{2023}\natexlab{}.
\newblock \showarticletitle{Diverse embedding expansion network and low-light cross-modality benchmark for visible-infrared person re-identification}. In \bibinfo{booktitle}{\emph{Proceedings of the IEEE/CVF conference on computer vision and pattern recognition}}. \bibinfo{pages}{2153--2162}.
\newblock


\bibitem[Zhang et~al\mbox{.}(2021)]%
        {mmn}
\bibfield{author}{\bibinfo{person}{Yukang Zhang}, \bibinfo{person}{Yan Yan}, \bibinfo{person}{Yang Lu}, {and} \bibinfo{person}{Hanzi Wang}.} \bibinfo{year}{2021}\natexlab{}.
\newblock \showarticletitle{Towards a unified middle modality learning for visible-infrared person re-identification}. In \bibinfo{booktitle}{\emph{Proceedings of the 29th ACM international conference on multimedia}}. \bibinfo{pages}{788--796}.
\newblock


\bibitem[Zhang et~al\mbox{.}(2022b)]%
        {msclnet}
\bibfield{author}{\bibinfo{person}{Yiyuan Zhang}, \bibinfo{person}{Sanyuan Zhao}, \bibinfo{person}{Yuhao Kang}, {and} \bibinfo{person}{Jianbing Shen}.} \bibinfo{year}{2022}\natexlab{b}.
\newblock \showarticletitle{Modality synergy complement learning with cascaded aggregation for visible-infrared person re-identification}. In \bibinfo{booktitle}{\emph{European conference on computer vision}}. Springer, \bibinfo{pages}{462--479}.
\newblock


\bibitem[Zhou et~al\mbox{.}(2022)]%
        {coop}
\bibfield{author}{\bibinfo{person}{Kaiyang Zhou}, \bibinfo{person}{Jingkang Yang}, \bibinfo{person}{Chen~Change Loy}, {and} \bibinfo{person}{Ziwei Liu}.} \bibinfo{year}{2022}\natexlab{}.
\newblock \showarticletitle{Learning to prompt for vision-language models}.
\newblock \bibinfo{journal}{\emph{International Journal of Computer Vision}} \bibinfo{volume}{130}, \bibinfo{number}{9} (\bibinfo{year}{2022}), \bibinfo{pages}{2337--2348}.
\newblock


\end{thebibliography}


\end{document}


\title{Prototypical Prompting for Text-to-image Person Re-identification}


\author{Shuanglin Yan}
\affiliation{%
  \institution{Nanjing University of Science and Technology}
  \state{Nanjing}
  \country{China}
}
\email{shuanglinyan@njust.edu.cn}

\author{Jun Liu}
\affiliation{%
  \institution{Singapore University of Technology and Design}
  \streetaddress{8 Somapah Rd}
  \country{Singapore}}
\email{jun_liu@sutd.edu.sg}

\author{Neng Dong}
\affiliation{%
  \institution{Nanjing University of Science and Technology}
  \state{Nanjing}
  \country{China}}
\email{neng.dong@njust.edu.cn}

\author{Liyan Zhang}
\affiliation{%
  \institution{Nanjing University of Aeronautics and Astronautics}
  \state{Nanjing}
  \country{China}}
\email{zhangliyan@nuaa.edu.cn}
  
\author{Jinhui Tang}
\affiliation{%
  \institution{Nanjing University of Science and Technology}
  \state{Nanjing}
  \country{China}}
\email{jinhuitang@njust.edu.cn}

\renewcommand{\shortauthors}{Shuanglin Yan, et al.}

\begin{abstract}
In this paper, we study the problem of Text-to-Image Person Re-identification (TIReID), which aims to find images of the same identity described by a text sentence from a pool of candidate images. Benefiting from Vision-Language Pre-training, such as CLIP (Contrastive Language-Image Pretraining), the TIReID techniques have achieved remarkable progress recently. However, most existing methods only focus on instance-level matching and ignore identity-level matching, that is, many-to-many matching between images and texts under the same identity. In this paper, we propose a novel prototypical prompting framework (Propot) designed to simultaneously model instance-level and identity-level matching for TIReID. Our Propot transforms the identity-level matching problem into a prototype learning problem, which aims to learn identity-enriched prototypes. Specifically, Propot works by ‘initialize, adapt, enrich, then aggregate’. We first utilize CLIP to produce initial prototypes to ensure their quality. Then, we propose a domain-conditional prototypical prompting (DPP) module to prompt the initial prototypes for adapting to the TIReID task through task-related information. Further, we propose an instance-conditional prototypical prompting (IPP) module to update prototypes conditioned on intra-modal and inter-modal instances to ensure prototype diversity. Finally, we design an adaptive prototype aggregation module to aggregate the above generated prototypes to generate the final identity-enriched prototypes. With identity-enriched prototypes, we diffuse the rich identity information of prototypes to instances through prototype-to-instance contrastive loss to model identity-level matching. Extensive experiments on three benchmarks demonstrate the superiority of Propot against existing TIReID methods.
\end{abstract}

\ccsdesc[100]{Information systems~Top-k retrieval in databases}

\keywords{Text-to-image person re-identification, Identity-level matching, Prototypical prompting}

\maketitle

\begin{figure}[!t]
\setlength{\abovecaptionskip}{0.1cm}
\setlength{\belowcaptionskip}{-0.4cm}
\subfigbottomskip=-2pt
\subfigcapskip=-2pt
\centering
\subfigure[Example of TIReID data]{\includegraphics[height=1.6in,width=3.2in]{images/motivation-a.pdf}}
\hspace{-10mm}
\subfigure[Existing matching paradigm] {\includegraphics[height=0.6in,width=3.2in]{images/motivation-b.pdf}}
\subfigure[Our proposed Propot] {\includegraphics[height=1.0in,width=3.2in]{images/motivation-c.pdf}}
\caption{The motivation of our proposed Propot. (a) Some examples of TIReID data containing multiple images from two identities and their annotated texts. Instances under the same identity are significantly different. (b) Most existing TIReID methods only focus on instance-level matching and ignore identity-level matching. (c) Our Propot proposes a prototype prompting framework to produce identity-enriched prototypes and diffuse their rich identity information to instances for modeling identity-level matching.}
\label{Fig:1}
\end{figure}

\section{Introduction}
Person re-identification (ReID), devoted to searching a person-of-interest across different times, locations, and camera views, has garnered increasing interest due to its huge practical value. In recent years, we have witnessed great progress on image-to-image person re-identification (IIReID)~\cite{li2023logical, AIESL, pifhsa, cdvr, GLMC}, which has been successfully applied in various practical scenarios. However, ReID is still challenging when pedestrian images from some cameras are missing. In contrast, textual descriptions are more readily accessible and freer than images collected by specialized equipment, which can be obtained from witnesses at the scene. Thus, text-to-image person re-identification (TIReID)~\cite{GNA} has received a lot of research attention recently owing to it being closer to real-world scenarios.

Compared with IIReID, the critical challenge of TIReID is to solve the modal gap between images and texts while modeling their correspondence. Various methods have been devised to address this challenge by cross-modal interactive attention mechanisms~\cite{HGAN, MIA, DSSL}, semantically aligned local feature learning~\cite{SSAN, tipcb, MANet, lgur}, and fine-grained auxiliary task-enhanced global feature learning~\cite{C2A2, ASAMN}. Nowadays, the emergence of vision-language pre-training models (VLP) has propelled the advancements in computer vision, showcasing exceptional capabilities in semantic understanding, multi-modal alignment, and generalization. It is intuitive to consider that the rich multi-modal prior information of VLP can be harnessed to enhance TIReID models in modeling inter-modal correspondence. Some follow-up CFine~\cite{CFine}, IRRA~\cite{irra} extend VLP for TIReID, greatly promoting the progress of this task. 

Despite the success of VLP-based methods, the current performance of TIReID lags behind that of IIReID, especially in complex scenes such as multiple cameras and drastic view variations. This is mainly attributed to the specificity of the TIReID task. As shown in Figure~\ref{Fig:1}(a), there are multiple images of the same identity, and each image is annotated with a specific text at the instance level. The purpose of TIReID is to find all images under the same identity given a text, that is, to correctly match each text with all images under the same identity. However, most existing methods limited by the alignment objective function~\cite{LCR2S} are trained to only consider \textbf{instance-level matching} between each image and its annotated text, ignoring matches with other non-annotated texts of the same identity, called \textbf{identity-level matching}. An intuitive idea to this problem is to directly match all images and texts under the same identity, but this crude strategy can disrupt instance-level matching relationship~\cite{LCR2S} due to huge view variations, leading to performance collapse. To date, there have been limited solutions~\cite{SSAN, LCR2S} proposed to model identity-level matching for TIReID. SSAN~\cite{SSAN} introduces an enhanced alignment loss to simultaneously align the image with its annotated text and another text sharing the same identity. LCR$^2$S~\cite{LCR2S} designs a teacher-student network architecture to randomly fuse two images/texts under the same identity and then align them to model identity-level matching. While effective, they do not comprehensively model the matching relationship between all images and texts under the same identity. Moreover, the two-stage teacher-student framework incurs high training costs and lacks practicality. In this paper, we thus ask: \emph{Can an end-to-end efficient TIReID model be trained to simultaneously model instance-level and identity-level matching?}

To address this problem, we propose a novel \textbf{Pro}totypical \textbf{p}r\textbf{o}mp-\\\textbf{t}ing framework (\textbf{Propot}) that enables the network to model instance-level and identity-level cross-modal matching for TIReID simultaneously. As shown in Figure \ref{Fig:1}, existing TIReID methods only model instance-level matching between each image and its labeled text. In contrast, our Propot framework introduces identity-level matching based on the above instance-level matching. We learn a prototype containing rich identity information for each identity, and diffuse the rich information contained in the prototype to each instance, indirectly modeling identity-level matching. This transforms the identity-level matching problem into an \textbf{identity-enriched prototype learning} problem. 

Specifically, our Propot follows the ‘initialize, adapt, enrich, then aggregate’ pipeline. Given the powerful multi-modal alignment capability of CLIP, we first cluster the instance (image/text) features extracted by CLIP under each identity as the initial prototype of each identity. Due to the domain gap between CLIP and TIReID training data, the initial prototype was not fully adapted to TIReID. To solve this problem, inspired by CoOp~\cite{coop}, we propose a domain-conditional prototypical prompting (DPP) module, which introduces a set of learnable prompt tokens to learn target domain knowledge and pass it to the initial prototype to adapt it to the TIReID task. However, instances under the same identity exhibit significant differences and diversity due to intrinsic factors such as view changes and camera parameters. If this diversity is ignored, the learned prototype will tend to be monotonous, and rich identity information will be lost. Therefore, inspired by CoCoOp~\cite{cocoop}, we propose an instance-condition prototypical prompting (IPP) module, which makes a prototypical prompt conditioned on a batch of instances. To further enrich the diversity of prototypes, we use both intra-modal instances and inter-modal instances to make two prototypical prompts, further narrowing the modal gap. Afterward, to fully integrate the multiple prototypes generated above, an adaptive prototype aggregation (APA) module is designed, which regards the initial prototype generated by CLIP as the prototype baseline and adaptively ensemble these generated prototypes as the final prototype. Finally, we utilize the prototype-to-instance contrastive loss to diffuse the rich identity information of the prototype to each instance, enabling the network to effectively model identity-level matching. Note that our Propot is single-stage and end-to-end trainable. During inference, only the backbone of the network is used for inference, which is simple and efficient. 

The main contributions of this paper are summarized as follows: (1) We propose an end-to-end trainable prototypical prompting framework to model instance-level and identity-level cross-modal matching for TIReID simultaneously. (2) We transform the identity-level matching problem into an identity-enriched prototype learning problem. Benefiting from CLIP and prompt learning, we utilize CLIP to generate initial prototypes and propose a domain-conditional prototypical prompting (DPP) module and an instance-condition prototypical prompting (IPP) module to generate identity-enriched prototypes. And an adaptive prototype aggregation (APA) module is designed to fully integrate the multiple prototypes. (3) Extensive experiments have been conducted to validate the effectiveness of Propot, and it achieves superior performance on the CUHK-PEDES, ICFG-PEDES, and RSTPReid benchmarks.

\vspace{-0.21cm}
\section{Related Work}
\subsection{Text-to-Image Person Re-identification}
TIReID~\cite{GNA} has been of growing interest over the last few years. Existing studies in TIReID can be broadly divided into the following categories: better model architectures and optimization losses, better alignment strategies, and richer prior information. Previous methods~\cite{Dual, CMPM, CMKA} have focused on designing network and optimization loss to learn globally aligned image and text features in a joint embedding space. These methods are simple and efficient, but ignore detailed information and fine-grained correspondences. To address the limitations of preceding approaches, subsequent methods have been dedicated to refining matching strategies to mine fine-grained correspondence between modalities. Some methods~\cite{GNA, HGAN, MIA, PMA, mgel, DSSL} have emerged to perform fine-grained matching through pairwise interactions between local parts of images and texts. While these approaches demonstrate superior performance by fostering ample cross-modal interactions, they come at the expense of escalating computational costs. To mitigate computational overhead, certain subsequent works~\cite{SSAN, tipcb, LapsCore} adopt local image parts as references to guide the generation of locally aligned text features, avoiding the pairwise interactions between them. However, the efficacy of these methods is constrained by the quality of explicitly acquired local parts. Consequently, alternative methods~\cite{MANet, saf, ivt, lgur} propose diverse aggregation schemes to adaptively aggregate images and text into a set of modality-shared local features, avoiding explicit local part acquisition. 


Recently, visual-language pre-training models (VLP)~\cite{clip} have made significant progress. To benefit from its rich multi-modal prior information, some recent state-of-the-art methods~\cite{TextReID, CFine, irra, empirical} have proposed diverse strategies to tailor VLP for TIReID, resulting in notable performance enhancements. Despite the substantial progress in TIReID, existing methods overlook the inherent nature of TIReID as an identity-level multi-view image-text matching challenge, distinct from a conventional instance-level image-text matching problem. Addressing this concern, LCR$^2$S~\cite{LCR2S} devised a teacher-student network to reason about a comprehensive representation containing multi-view information from a single image or text. This approach yielded significantly improved performance, albeit with limitations in fully integrating multi-view information and exhibiting lower training efficiency. In this study, we present an end-to-end framework to learn a prototype with comprehensive information for each identity based on the entire training set, and transfer multi-view information to individual samples for identity-level matching, providing a solution that enhances both efficiency and the integration of multi-view data for TIReID.

\vspace{-4pt}
\subsection{Vision-Language Pre-Training}
Nowadays, the "pre-training and fine-tuning" paradigm stands as a foundational approach in the computer vision community, in which pre-training models that can provide rich prior knowledge for various downstream vision tasks have gained increasing attention. Previous prevailing practice is rooted in the supervised unimodal pre-training~\cite{vit, resnet} on ImageNet~\cite{imagenet}, followed by fine-tuning on various downstream tasks. In response to the need for enhanced representation capabilities and to overcome annotation constraints, a novel paradigm known as language-supervised vision pre-training (vision-language pre-training, VLP) has emerged. Within VLP, investigating the interaction between vision and language has become a central research focus. Several works~\cite{vilbert, visualbert, uniter, albef, beit} have been proposed to model the interaction of vision and language based on some multi-modal reasoning tasks, such as masked language/region modeling, and image captioning. Recently, contrastive representation learning~\cite{clip, align, filip}, which learns representations by contrasting positive pairs against negative pairs, has gotten well studied for diverse visual representation learning. The representative work, Contrastive Language-Image Pretraining (CLIP)~\cite{clip}, has strong multi-modal semantic representation and zero-shot generalization capabilities, which is trained on 400 million image-text pairs. CLIP has shown promising adaptability for various downstream tasks,  such as video-text retrieval~\cite{clip4clip, clip2video}, referring image segmentation~\cite{cris}, and person re-identification~\cite{CFine}. Our work is built upon the CLIP and leverages its ample multi-modal prior information for learning identity-enriched prototypes.

\vspace{-4pt}
\subsection{Prompt Learning}
Prompt learning~\cite{autoprompt, how} originates from the natural language processing (NLP) domain, which aims to adapt pre-training models to various downstream tasks through prompting. Prompts are typically instructions in the form of sentences that are given to the pre-training model to better understand the downstream task. Early prompts were manually designed for a specific task to better understand the task. However, due to issues such as instability and knowledge bias, recent studies have introduced the concept of prompt learning, wherein task-adapted prompts are automatically generated during the fine-tuning stage. Currently, prompt learning has been extended to the computer vision domain. CoOp~\cite{coop} pioneered the application of prompt learning to adapt large vision-language models in computer vision, while CoCoOp~\cite{cocoop} built upon CoOp to introduce a conditional prompt learning framework, addressing challenges related to weak generalization. Chen et al.~\cite{pvu} proposed to efficiently adapt the CLIP model to the resource-hungry video understanding task by optimizing a few continuous prompt vectors. CLIP-ReID~\cite{clipreid} designed a two-stage framework to generate coarse descriptions of pedestrians to fully utilize the powerful capabilities of CLIP for the ReID task. Inspired by these, in this work we propose to leverage prompt learning to learn comprehensive prototype representations to model many-to-many matching for TIReID.

\begin{figure*}[t!]
  \centering
  \setlength{\abovecaptionskip}{0.1cm}
  \setlength{\belowcaptionskip}{-0.2cm}
  \includegraphics[width=\linewidth,height=3.8in]{images/framework.pdf}\\
  \caption{Overview of our Propot. It includes instance-level matching and identity-enriched prototype learning. For instance-level matching, each image and its annotated text are directly aligned through SDM loss (Baseline). For prototype learning, we first utilize pre-trained CLIP to generate the initial prototypes ($\bm {pt}^v$ and $\bm {pt}^t$). We then adapt the initial prototypes to TIReID through the DPP module to generate the task-adapted prototypes ($\bm {p}_a^v$ and $\bm {p}_a^t$). And the IPP module updates the prototypes conditioned on a batch of intra-modal and inter-modal instances to generate intra-modal and inter-modal enriched prototypes ($\bm {p}_{en}^v$, $\bm {p}_{en}^t$, $\bm {p}_{eo}^v$ and $\bm {p}_{eo}^t$), respectively. The above multiple prototypes are aggregated through Adaptive Prototypical Aggregation (APA) to generate the final prototypes ($\bm {p}^v$ and $\bm {p}^t$), and their rich identity information is diffused to each instance through prototype-to-instance contrastive loss ($\mathcal{L}_{p2v}$, $\mathcal{L}_{p2t}$) to model identity-level matching. Moreover, we also introduce the MLM module as compensation to model fine-grained matching. During testing, only visual and textual encoders are used for inference.}
  \label{Fig:2}
\end{figure*}

\section{The Propot Framework}
Our Propot is a conceptually simple end-to-end trainable framework, and the overview is depicted in Figure~\ref{Fig:2}. We first extract features of each image and text instance through visual and textual encoders, followed by instance-level and identity-level matching. For instance-level matching, we follow existing matching paradigms by directly minimizing the distance between each image and its annotated text. For identity-level matching, we transform it into an identity-enriched prototype learning problem. The aim is to learn a rich prototype containing all instance identity information for each identity in the training set. 
Further details on Propot are expounded in the subsequent subsections.

\subsection{Feature Extraction}
Prior studies~\cite{CFine, irra, empirical} have underscored the effectiveness of CLIP~\cite{clip} in addressing the challenges of TIReID. To harness the extensive multi-modal prior knowledge embedded in CLIP, we leverage CLIP’s image and text encoders to initialize Propot’s image and text backbones. Concretely, given an image-text pair $(I, T)$, we exploit CLIP pre-trained ViT model for extracting visual representations for the image $I$, resulting in the global visual feature $\bm v \in \mathbb{R}^{d}$. For the text caption $T$, the CLIP's textual encoder is utilized to generate the global textual feature $\bm t \in \mathbb{R}^{d}$.


The instance-level matching is modeled by directly aligning $\bm v$ and $\bm t$. As previously highlighted, to model identity-level matching, it becomes imperative to generate a prototype containing rich identity information for each identity. In Propot, the prototype learning method plays a pivotal role. While it is conceivable to learn a prototype for each identity from scratch, akin to previous methodologies~\cite{clipreid}, such an approach introduces challenges in network convergence and does not guarantee prototype quality. To address these concerns, we introduce a novel ‘initialize, adapt, enrich, then aggregate’ prototype learning scheme, detailed in the subsequent subsections.

\subsection{Initial Prototype Generation} 
Leveraging the powerful semantic information extraction and multi-modal alignment capabilities inherent in the CLIP model, we initiate our approach by employing CLIP to extract features for each instance within the TIReID training set. Subsequently, we cluster these instance features based on shared identity labels to produce initial prototypes for the respective modalities. Specifically, given a TIReID training set $\{I_i, T_i, Y_i\}_{i=1}^{N_s}$, where $Y_i \in\{L_1, L_2, ..., L_N\}$ represents the identity label of the image-text pair $(I_i, T_i)$, $N_s$ denotes the number of image-text pairs, and $N$ denotes the number of identity label, we employ the pre-trained CLIP visual and textual encoders to obtain the visual and textual features $\{\bm v_{i}, \bm t_{i}\}_{i=1}^{N_s}$ of all image-text pairs in the training set. We then perform feature aggregation on the image and text features. Taking identity $L_i$ as an example, We generate initial prototypes $\bm {pt}_i^v$ and $\bm {pt}_i^t$ for identity $L_i$ as follows:
\begin{equation}\
\begin{aligned}
\bm {pt}_i^v=\sum_{j=1,Y_j \in L_i}^{N_i}\bm v_{j}, \quad \bm {pt}_i^t=\sum_{j=1,Y_j \in L_i}^{N_i}\bm t_{j}
\end{aligned},
\end{equation}

where $N_i$ denotes the number of instances under the identity $L_i$. Therefore, for the entire training set, we can generate visual and textual initial prototype sets $\bm {pt}^v=[\bm {pt}_1^v, \bm {pt}_2^v, ..., \bm {pt}_N^v]\in \mathbb{R}^{N\times d}$ and $\bm {pt}^t=[\bm {pt}_1^t, \bm {pt}_2^t, ..., \bm {pt}_N^t]\in \mathbb{R}^{N\times d}$.

\subsection{Domain-conditional Prototypical Prompting}
Capitalizing on the potent capabilities of CLIP, the initial prototype effectively captures certain identity information to model identity-level matching, leading to improved performance in TIReID (as shown in Table IV), which also confirms the viability of our idea. However, a notable challenge arises due to the domain gap between the pre-training data of CLIP and the TIReID dataset. Consequently, the identity information mined by CLIP's features falls short, significantly diminishing the impact of the initial prototype. In response to this challenge and drawing inspiration from the Contextual Optimization (CoOp) framework~\cite{coop}, which introduces a set of learnable context vectors to adapt VLMs to downstream tasks, we introduce a Domain-conditional Prototypical Prompting (DPP) module to adapt the initial prototype to the TIReID task. Specifically, we incorporate a set of learnable contextual prompt vectors preceding each initial prototype. These vectors undergo training on the TIReID dataset concurrently with the network, acquiring domain knowledge specific to the TIReID task. These vectors then transmit the domain knowledge to each initial prototype through the self-attention encoder (SAE), facilitating the adaptation of prototypes to the TIReID task.

Formally, for each initial prototype $\bm {pt}_i$, we add a set of learnable contextual prompt vectors $\{[\bm X_i]_1, [\bm X_i]_2, ..., [\bm X_i]_K\}\in \mathbb{R}^{K\times d}$ in front of it, where $[\bm X_i]$ is the visual contextual prompt vector $\bm c_{v,i}$ for the visual prototype $\bm {pt}_i^v$, and $[\bm X_i]$ is the textual contextual prompt vector $\bm c_{t,i}$ for the textual prototype $\bm {pt}_i^t$. Then, we feed $\{[\bm X_i]_1, [\bm X_i]_2, ..., [\bm X_i]_K, \bm {pt}_i\}\in \mathbb{R}^{(K+1)\times d}$ into the self-attention encoder (SAE) to pass the information to the initial prototype for updating it.
\begin{equation}\
\begin{aligned}
\bm p_{a,i}=SAE(\{[\bm X_i]_1, [\bm X_i]_2, ..., [\bm X_i]_K, \bm {pt}_i\})
\end{aligned},
\end{equation}
where $\bm p_{a,i}\in \mathbb{R}^{d}$ represents the task-adaptive prototype. SAE is comprised of $N_a$ blocks, with each block containing a multi-head self-attention layer and a feed-forward network layer. The above process allows us to generate modality-specific task-adaptive prototypes, denoted as $\bm p^v_{a,i}$ and $\bm p^t_{a,i}$, respectively, according to the input prototypes of different modalities.

\subsection{Instance-conditional Prototypical Prompting}
So far, we have generated a prototype adapted to the TIReID task. As expected, the resulting prototype resulted in significant performance improvements. However, as depicted in Figure 1, instances under the same identity showcase notable differences and diversity attributable to intrinsic factors like view changes and camera parameters. The previously derived prototype fails to account for this diversity, resulting in the loss of certain discriminative identity information. To solve this problem, inspired by Conditional Context Optimization (CoCoOp)~\cite{cocoop}, we propose an Instance-conditional Prototypical Prompting (IPP) module. This module serves to update the prototype conditioned on each batch of the input instances, allowing for the comprehensive mining of instance-specific information and ensuring diversity within the prototype.

Specifically, for a batch of $B$ image-text pairs $\{I_i, T_i\}_{i=1}^B$, after feature encoding, the encoded features can be expressed as $\bm V_B=[\bm v_{1}, \bm v_{2}, ..., \bm v_{B}]\in \mathbb{R}^{B\times d}$ and $\bm T_B=[\bm t_{1}, \bm t_{2}, ..., \bm t_{B}]\in \mathbb{R}^{B\times d}$. Subsequently, the instance features $\bm V_B/\bm T_B$ and the prototypes $\bm {pt}^v/\bm {pt}^t$ undergo processing in a cross-attention decoder (CAD). This decoder facilitates the interaction between instance features and prototypes, allowing for the extraction of identity-relevant information from instances to enrich the prototypes. There are $N_e$ blocks in CAD, each featuring a multi-head cross-attention layer and a feed-forward network layer. For the cross-attention decoder, the prototypes $\bm {pt}^v/\bm {pt}^t$ serve as \emph{query}, while the instance features $\bm V_B/\bm T_B$ are designated as \emph{key} and \emph{value}, which is formulated as follows:
\begin{equation}\
\begin{aligned}
\bm p_{en,i}^v=CAD(\bm {pt}^v, \bm V_B, \bm V_B),\quad\bm p_{en,i}^t=CAD(\bm {pt}^t, \bm T_B, \bm T_B)
\end{aligned},
\end{equation}
where $\bm p_{en,i}^v\in \mathbb{R}^{d}$ and $\bm p_{en,i}^t\in \mathbb{R}^{d}$ represent the intra-modal instance-enriched visual and textual prototypes, respectively. The above leverages the intra-modal instance features to enrich the prototypes. Additionally, we extend this enrichment by incorporating inter-modal instance features, fostering the network to extract modality-shared identity information comprehensively and minimize the modal gap. We formulate them as
\begin{equation}\
\begin{aligned}
\bm p_{eo,i}^v=CAD(\bm {pt}^v, \bm T_B, \bm T_B),\quad \bm p_{eo,i}^t=CAD(\bm {pt}^t, \bm V_B, \bm V_B)
\end{aligned},
\end{equation}
where $\bm p_{eo,i}^v\in \mathbb{R}^{d}$ and $\bm p_{eo,i}^t\in \mathbb{R}^{d}$ represent the inter-modal instance-enriched visual and textual prototypes, respectively.

\subsection{Adaptive Prototype Aggregation}
Consequently, for both images and texts, we generate three distinct modality-specific prototypes: $\bm p_{a,i}^m\in \mathbb{R}^{d}$, $\bm p_{en,i}^m\in \mathbb{R}^{d}$, and $\bm p_{eo,i}^m\in \mathbb{R}^{d}$, each containing different information for each identity $L_i$, where $m\in \{v, t\}$. To seamlessly integrate all available information, we introduce an Adaptive Prototype Aggregation (APA) module. This module effectively aggregates diverse prototypes to construct a comprehensive identity-enriched prototype. As $\bm {pt}_i^m$ is directly derived by CLIP, benefiting from its strong semantic understanding capabilities, the quality of the prototype is assured. We designate $\bm {pt}i^m$ as the prototype baseline for the aggregation process and the weights for aggregation are determined by the correlation of other prototypes with $\bm {pt}i^m$. This approach enables the suppression of spurious identity prototype information in $\bm p{a,i}^m, \bm p{en,i}^m, \bm p_{eo,i}^m$, while amplifying the correct ones during aggregation. The calculation of prototype correlation, serving as the aggregation weights for the three prototypes, is articulated as follows: 
\begin{equation}\
\begin{aligned}
\bm w_{a,i}^m=\bm {pt}_i^m(\bm p_{a,i}^m)^T,\quad\bm w_{en,i}^m=\bm {pt}_i^m(\bm p_{en,i}^m)^T,\quad\bm w_{en,i}^m=\bm {pt}_i^m(\bm p_{en,i}^m)^T
\end{aligned}.
\end{equation}

Then, we utilize the softmax function to normalize the weights and obtain the final identity-enriched prototype as
\begin{equation}\
\begin{aligned}
\bm p_i^m=\bm {pt}_i^m+\sum_{k}\bm w_{k,i}^m\cdot\bm p_{k,i}^m
\end{aligned},
\end{equation}
where $k\in \{a, en, eo\}$. Thus, for the entire training set, we can generate the final visual and textual prototype sets $\bm {p}^v=[\bm {p}_1^v, \bm {p}_2^v, ..., \bm {p}_N^v]\in \mathbb{R}^{N\times d}$ and $\bm {p}^t=[\bm {p}_1^t, \bm {p}_2^t, ..., \bm {p}_N^t]\in \mathbb{R}^{N\times d}$. Subsequently, we propagate the rich identity information encapsulated in the prototypes to each instance through prototype-to-instance contrastive loss to model identity-level matching for TIReID.

\subsection{Training and Inference}
The goal of Propot is to model both instance-level and identity-level matching for TIReID. To this end, we optimize Propot through cross-modal matching loss, cross-entropy loss, prototype-to-instance contrastive loss, and mask language modeling loss. 

Given a batch of $B$ image-text pairs $\{I_i, T_i\}_{i=1}^B$, we generate the global visual and textual features as $\bm V_B=[\bm v_{1}, \bm v_{2}, ..., \bm v_{B}]\in \mathbb{R}^{B\times d}$ and $\bm T_B=[\bm t_{1}, \bm t_{2}, ..., \bm t_{B}]\in \mathbb{R}^{B\times d}$. To align each image $I_i$ and its annotated text $T_i$, we utilize the similarity distribution matching (SDM)~\cite{irra} as the cross-modal matching loss to model instance-level matching between them.
\begin{equation}\
\begin{aligned}
\mathcal{L}_{sdm}=\mathcal{L}_{i2t}+\mathcal{L}_{t2i}
\end{aligned},
\end{equation}
\begin{equation}\
\begin{aligned}\label{Eq:13}
\mathcal{L}_{i2t}=\frac{1}{B}\sum_{i=1}^{B}\sum_{j=1}^{B}p_{i,j}log(\frac{p_{i,j}}{q_{i,j}+\epsilon})
\end{aligned},
\end{equation}
\begin{equation}\
\begin{aligned}\label{Eq:14}
p_{i,j}=\frac{exp(sim(\bm v_i, \bm t_j)/\tau)}{\sum_{k=1}^Bexp(sim(\bm v_i, \bm t_k)/\tau)}
\end{aligned},
\end{equation}
where $sim(\cdot)$ denotes the cosine similarity function, $\tau$ denotes the temperature factor to control the distribution peaks, and $\epsilon$ is a small number to avoid numerical problems. $q_{i,j}$ denotes the true matching probability. $\alpha$ indicates the margin. $\mathcal{L}_{t2i}$ can be obtained by exchanging $\bm v$ and $\bm t$ in Eqs.~\ref{Eq:13} and \ref{Eq:14}. Moreover, to ensure the discriminability of features $\bm v$ and $\bm t$, we calculate cross-entropy loss $\mathcal{L}_{id}$ on them to classify them into corresponding identity labels.

Through the prototype learning process described above, we generate two identity-rich prototypes  $\bm {p}_i^v$ and $\bm {p}_i^t$ for each identity $L_i$. To effectively diffuse the rich identity information encapsulated in the prototypes to instances of the same identity, we employ a prototype-to-instance contrastive loss, denoted as $\mathcal{L}_{p2i}$. This loss operates in tandem with the cross-modal matching loss, collectively contributing to the modeling of identity-level matching.
\begin{equation}\
\begin{aligned}
\mathcal{L}_{p2i}=\sum_{i=1}^{N}\mathcal{L}_{p2v}(L_i)+\mathcal{L}_{p2t}(L_i)
\end{aligned},
\end{equation}
\begin{equation}\
\begin{aligned}
\mathcal{L}_{p2v}(L_i)=-\frac{1}{|P(L_i)|}\sum_{p\in P(L_i)}\frac{exp(sim(\bm v_p, \bm p_i^v)/\tau}{\sum_{k=1}^Bexp(sim(\bm v_j, \bm p_i^v)/\tau)}
\end{aligned},
\end{equation}
\begin{equation}\
\begin{aligned}
\mathcal{L}_{p2t}(L_i)=-\frac{1}{|P(L_i)|}\sum_{p\in P(L_i)}\frac{exp(sim(\bm t_p, \bm p_i^t)/\tau}{\sum_{k=1}^Bexp(sim(\bm t_j, \bm p_i^t)/\tau)}
\end{aligned},
\end{equation}
where $P(L_i)$ represents a set of instance indices of identity $L_i$, and $|P(L_i)|$ denotes the cardinality of $P(L_i)$. To further improve TIReID performance, we follow~\cite{irra} to introduce a mask language modeling task denoted by $\mathcal{L}_{mlm}$ to model fine-grained matching between modalities.

Propot is a single-stage and end-to-end trainable framework, and the overall objective function $\mathcal{L}$ for training is as follows:
\begin{equation}\
\begin{aligned}
\mathcal{L}=\mathcal{L}_{sdm}+\mathcal{L}_{id}+\lambda_1\mathcal{L}_{p2i}+\lambda_2\mathcal{L}_{mlm}
\end{aligned},
\end{equation}
where $\lambda_1$ and $\lambda_2$ balance the contribution of different loss terms during training.

After training, Propot has the ability of instance-level and identity-level matching. The prototype learning process only exists during the training phase. In the inference phase, we only use Propot’s visual and textual encoders to extract the global features of the test samples. The retrieval results are then obtained by calculating the cosine similarity. 

\section{Experiments}
\subsection{Experiment Settings}
\textbf{Datasets and Metrics:}
The evaluations are implemented on three benchmark datasets for TIReID, namely CUHK-PEDES~\cite{GNA}, ICFG-PEDES~\cite{SSAN}, and RSTPReid~\cite{DSSL}. \textbf{CUHK-PEDES} comprises 40,206 images and 80,412 descriptions of 13,003 persons. Each image is annotated with 2 descriptions, and the average length of each description is not less than 23 words. Following the dataset split of~\cite{GNA} for equitable comparisons, the training set encompasses 34,054 images and 68,108 descriptions of 11,003 persons, the validation set includes 3,078 images and 6,156 descriptions of 1,000 persons, and the testing set involves 3,074 images and 6,148 descriptions of 1,000 persons. \textbf{ICFG-PEDES} comprises 54,522 image-text pairs of 4,102 persons, with an average description length of 37 words. We adhere to the dataset split strategy outlined in~\cite{SSAN}, utilizing 34,674 image-text pairs of 3,102 persons for model training and reserving the remaining 1,000 persons for evaluation, totaling 19,848 image-text pairs. \textbf{RSTPReid} includes 20,505 images of 4,101 persons, each annotated with 2 descriptions. Each person has 5 images captured by multiple cameras. The average length of each description is not less than 23 words. There are 3,701 persons in the training set, 200 persons in the validation set, and 200 persons in the testing set. 

For performance evaluation, we employ the cumulative matching characteristics curve, a.k.a, Rank-$k$ matching accuracy (R@$k$). This metric signifies the probability that a correct match appears within the top-$k$ ranked retrieved results. Consistent with convention, we report matching accuracy for $k$=${1, 5, 10}$.

\begin{table}[!ht]\small
\setlength{\abovecaptionskip}{0.1cm} 
\setlength{\belowcaptionskip}{-0.2cm}
\centering {\caption{Performance comparison with state-of-the-art methods on CUHK-PEDES. R@1, R@5, and R@10 are listed.}\label{Tab:1}
\renewcommand\arraystretch{1.0}
\begin{tabular}{c|c|c|ccc}
 \hline
  Methods & Pre &Ref & R@1 & R@5 & R@10\\
  \hline  
  





  
  
  


  ACSA~\cite{ACSA} &\multirow{18}{*}{\rotatebox{90}{w/o CLIP}}   & TMM'22  & 63.56   & 81.40  & 87.70 \\
  

  SRCF~\cite{SRCF} &  & ECCV'22 & 64.04   & 82.99   & 88.81 \\

  LBUL~\cite{LBUL} &  & MM'22 & 64.04  & 82.66  & 87.22    \\
  
  
  CAIBC~\cite{caibc} &    & MM'22 & 64.43  & 82.87   & 88.37 \\
  
  AXM-Net~\cite{AXM-Net}  &   & AAAI'22 & 64.44  &80.52  &86.77 \\

  C$_2$A$_2$~\cite{C2A2}  &    & MM'22 & 64.82  & 83.54  & 89.77 \\ 

  LGUR~\cite{lgur}  &    & MM'22  & 65.25  & 83.12  & 89.00 \\

  FedSH~\cite{fedsh} &    & TMM'23 & 60.87 &  80.82 &  87.61 \\
  
  RKT~\cite{RKT} &    & TMM'23 & 61.48  &80.74  &87.28 \\


  PBSL~\cite{pbsl} & & MM'23  & 65.32 & 83.81 & 89.26  \\

  BEAT~\cite{beat} & & MM'23  & 65.61  & 83.45 & 89.57  \\

  MANet~\cite{MANet}  &  & TNNLS'23 & 65.64 & 83.01 & 88.78  \\

  ASAMN~\cite{ASAMN} & & TIP'23  & 65.66 & 84.53 & 90.21   \\

  MSN-BRR~\cite{MSN-BRR} & & ICMR'23  & 65.93 & 83.94 & 90.15  \\

  BDNet~\cite{bdnet} &    & PR'23 & 66.27 & 85.07 & 90.27 \\

  LCR$^2$S~\cite{LCR2S}  &  & MM'23 & 67.36  & 84.19  & 89.62 \\

  TransTPS~\cite{TransTPS} & & TMM'23  & 68.23 & 86.37 & 91.65  \\

  MGCN~\cite{mgcn} & & TMM'23  & 69.40 & 87.07 & 90.82  \\
  \hline

  CFine~\cite{CFine} &\multirow{14}{*}{\rotatebox{90}{w/ CLIP}}   & TIP'23 & 69.57 & 85.93 & 91.15  \\

  Sew~\cite{sew} &    & arXiv'23 & 69.61 & 86.01  & 90.90 \\

  CSKT~\cite{CSKT} &    & arXiv'23 & 69.70 & 86.92 & 91.8 \\

  VLP-TPS~\cite{VLP-TPS}  &  & arXiv'23 & 70.16 & 86.10 & 90.98 \\

  VGSG~\cite{vgsg}  &  & TIP'23 & 71.38 & 86.75 & 91.86 \\

  IRRA~\cite{irra}  &    & CVPR'23 & 73.38 & 89.93 & 93.71 \\

  BiLMa~\cite{BiLMa}  &  & ICCVW'23 & 74.03 & 89.59 & 93.62 \\

  TCB~\cite{TCB}  &  & MM'23 & 74.45 & 90.07 & 94.66  \\
  

  DCEL~\cite{DCEL} &  & MM'23 & 75.02 & 90.89 & 94.52  \\

  SAL~\cite{sal} &  & MMM'24 & 69.14 & 85.90 & 90.81 \\

  EESSO~\cite{eesso} &  & IVC'24 & 69.57 & 85.65 & 90.71 \\
  
  PD~\cite{pd} &  & arXiv'24 & 71.59 & 87.95 & 92.45 \\

  CFAM~\cite{CFAM}  &  & CVPR'24 & 72.87 & 88.61 & 92.87 \\

  TBPS-CLIP~\cite{empirical} &  & AAAI'24 & 73.54 & 88.19 & 92.35 \\
  \hline\hline
  \textbf{Ours} &   & - & \textbf{74.89} & \textbf{89.90} & \textbf{94.17}  \\
  \hline
\end{tabular}}
\end{table}

\textbf{Implementation Details:}
For input images, we uniformly resize them to 384$\times$128 and augment them with random horizontal flipping, random crop with padding, and random erasing. For input texts, the length of the textual token sequence is unified to 77, and we also augment texts by randomly masking out some text tokens with a probability of 15\% and replacing them with the special token $[MASK]$~\cite{bert}. Propot's visual encoder is initialized with the CLIP-ViT-B/6 version of CLIP, where the dimension $d$ of global visual and textual features is set to 512. The Self-Attention Encoder (SAE) and Cross-Attention Decoder (CAD) modules consist of $N_a=1$ block and $N_e=3$ blocks, respectively, each with 8 heads. The length $K$ of the learnable contextual prompt vector in the DPP module is set to 4. In both the SDM and prototype-to-instance contrastive losses, the temperature factor $\tau$ is set to 0.02. The loss balance factors are set to $\lambda_1=0.2$ and $\lambda_2=1.0$. Model training utilizes the Adam optimizer with a weight decay factor of 4e-5. Initial learning rates are set to 1e-5 for the visual/textual encoder and 1e-4 for other network modules. The learning rate is decreased using a cosine learning rate decay strategy, and the training is stopped at 60 epochs. Additionally, the learning rate linearly decays by a factor of 0.1 within the first 10\% of the training epochs for warmup. All experiments are implemented in PyTorch library, and models are trained with a batch size of 64 on a single RTX3090 24GB GPU. 

\subsection{Comparisons with State-of-the-art Models}
In this section, we compare our Propot with current state-of-the-art approaches on all the above three TIReID benchmarks. The methods for comparison are categorized into two sections. The first section includes methods (w/o CLIP) that initialize the backbone solely using single-modal pre-training models (ResNet~\cite{resnet}, ViT~\cite{vit}, BERT~\cite{bert}). In the second section, we present methods (w/ CLIP) that utilize the multi-modal pre-training CLIP~\cite{clip} model to initialize the backbone. Propot falls under the second section. 

\textbf{Evaluation on CUHK-PEDES:}
The performance comparison with state-of-the-art methods on the CUHK-PEDES dataset is summarized in Table \ref{Tab:1}. The proposed Propot framework demonstrates competitive performance at all metrics and outperforms all compared methods except~\cite{DCEL}, achieving remarkable R@1, R@5, and R@10 accuracies of 74.89\%, 89.90\% and 94.17\%, respectively. While our R@1 accuracy is slightly lower (-0.13\%) compared to the optimal method DCEL~\cite{DCEL}, it is essential to note that DECL introduces both mask language modeling and global-local semantic alignment to mine fine-grained matching, resulting in higher computational cost. In contrast, Propot employs only mask language modeling for fine-grained matching. Additionally, as observed in Table~\ref{Tab:4}\#7, even without a local matching module, Propot achieves a noteworthy 74.37\% R@1 accuracy, surpassing most compared methods. This is attributed to the fact that our prototypical prompting can simultaneously model instance-level and identity-level matching. 

\begin{table}[!t]\small
\setlength{\abovecaptionskip}{0.1cm} 
\setlength{\belowcaptionskip}{-0.2cm}
\centering {\caption{Performance comparison with state-of-the-art methods on ICFG-PEDES. R@1, R@5, and R@10 are listed.}\label{Tab:2}
\renewcommand\arraystretch{1.0}
\begin{tabular}{c|c|c|ccc}
\hline
  Methods & Pre & Ref & R@1 & R@5 & R@10 \\
  \hline  


  
  IVT~\cite{ivt} &\multirow{12}{*}{\rotatebox{90}{w/o CLIP}}   & ECCVW'22 & 56.04  & 73.60  & 80.22 \\

  SRCF~\cite{SRCF} & & ECCV'22 & 57.18   & 75.01  & 81.49 \\

  LGUR~\cite{lgur} & & MM'22  & 57.42 & 74.97  & 81.45 \\

  FedSH~\cite{fedsh} &    & TMM'23 & 55.01  & 72.75 & 79.48 \\

  ASAMN~\cite{ASAMN} & & TIP'23  & 57.09 & 76.33 & 82.84   \\


  BDNet~\cite{bdnet} &    & PR'23 & 57.31 & 76.15 & 81.58 \\

  PBSL~\cite{pbsl} & & MM'23  & 57.84 & 75.46 & 82.15  \\

  LCR$^2$S~\cite{LCR2S}  &  & MM'23 & 57.93  & 76.08  & 82.40 \\

  BEAT~\cite{beat} & & MM'23  & 58.25  & 75.92 & 81.96  \\

  MSN-BRR~\cite{MSN-BRR} & & ICMR'23  & 58.63 & 75.78 & 82.15 \\

  MANet~\cite{MANet} &  & TNNLS'23 & 59.44 & 76.80 & 82.75 \\

  MGCN~\cite{mgcn} & & TMM'23  & 60.20 & 76.75 & 83.90  \\
  \hline
  CSKT~\cite{CSKT} & \multirow{12}{*}{\rotatebox{90}{w/ CLIP}}  & arXiv'23 & 58.90 & 77.31 & 83.56 \\
  
  VLP-TPS~\cite{VLP-TPS}  &  & arXiv'23 & 60.64 & 75.97 & 81.76 \\

  CFine~\cite{CFine}  &   & TIP'23 & 60.83 & 76.55 & 82.42  \\

  TCB~\cite{TCB}  &  & MM'23 & 61.60 & 76.33 & 81.90  \\
  
  Sew~\cite{sew}  & & arXiv'23 & 62.29 & 77.15  &  82.52 \\

  VGSG~\cite{vgsg}  &  & TIP'23 & 63.05 & 78.43 & 84.36 \\

  IRRA~\cite{irra}  &    & CVPR'23 & 63.46  & 80.25 &  85.82 \\

  BiLMa~\cite{BiLMa}  &  & ICCVW'23 & 63.83 & 80.15 & 85.74 \\


  DCEL~\cite{DCEL} &  & MM'23 & 64.88 & 81.34 & 86.72  \\

  EESSO~\cite{eesso} &  & IVC'24 & 60.84 & 77.89 & 83.53 \\

  PD~\cite{pd} &  & arXiv'24 & 60.93 & 77.96 & 84.11 \\

  CFAM~\cite{CFAM}  &  & CVPR'24 & 62.17 & 79.57 & 85.32 \\

  SAL~\cite{sal} &  & MMM'24 & 62.77 & 78.64 & 84.21 \\

  TBPS-CLIP~\cite{empirical} &  & AAAI'24 & 65.05 & 80.34 & 85.47 \\
  \hline\hline
  \textbf{Ours} &  & - & \textbf{65.12} & \textbf{81.57}  & \textbf{86.97}   \\
  \hline
\end{tabular}}
\end{table}

\begin{table}[!t]\small
\setlength{\abovecaptionskip}{0.1cm} 
\setlength{\belowcaptionskip}{-0.2cm}
\centering {\caption{Performance comparison with state-of-the-art methods on RSTPReid. R@1, R@5, and R@10 are listed.}\label{Tab:3}
\renewcommand\arraystretch{1.0}
\begin{tabular}{c|c|c|ccc}
 \hline
  Methods & Pre & Ref & R@1 & R@5 & R@10 \\
  \hline



  LBUL~\cite{LBUL} &\multirow{9}{*}{\rotatebox{90}{w/o CLIP}}  & MM'22   & 45.55  & 68.20 & 77.85  \\
  
  IVT~\cite{ivt}  &  & ECCVW'22 & 46.70  & 70.00  & 78.80 \\

  ACSA~\cite{ACSA} &  & TMM'22  & 48.40  & 71.85  & 81.45 \\
  
  C$_2$A$_2$~\cite{C2A2}  &  & MM'22 & 51.55  & 76.75 & 85.15 \\

  PBSL~\cite{pbsl} & & MM'23  & 47.80 & 71.40 & 79.90  \\

  BEAT~\cite{beat} & & MM'23  & 48.10 & 73.10  & 81.30 \\


  MGCN~\cite{mgcn} & & TMM'23  & 52.95 & 75.30 & 84.04  \\

  LCR$^2$S~\cite{LCR2S}  & & MM'23 & 54.95  & 76.65  & 84.70 \\

  TransTPS~\cite{TransTPS} & & TMM'23  & 56.05 & 78.65 & 86.75  \\
  \hline
  CFine~\cite{CFine}  &\multirow{12}{*}{\rotatebox{90}{w/ CLIP}}  & TIP'23 & 50.55 & 72.50 & 81.60  \\
  
  VLP-TPS~\cite{VLP-TPS}  &  & arXiv'23 & 50.65 & 72.45 & 81.20 \\
  
  Sew~\cite{sew} &  & arXiv'23 & 51.95  & 73.50 & 82.45 \\

  CSKT~\cite{CSKT} &   & arXiv'23 & 57.75 & 81.30 & 88.35 \\

  IRRA~\cite{irra}  &    & CVPR'23 & 60.20  &  81.30 &  88.20 \\

  BiLMa~\cite{BiLMa}  &  & ICCVW'23 & 61.20  & 81.50  & 88.80 \\

  DCEL~\cite{DCEL} &  & MM'23 & 61.35 & 83.95 & 90.45  \\


  EESSO~\cite{eesso} &  & IVC'24 & 53.15 & 74.80 & 83.55 \\
  
  PD~\cite{pd} &  & arXiv'24 & 56.65 & 77.40 & 84.70 \\

  CFAM~\cite{CFAM}  &  & CVPR'24 & 59.40 & 81.35 & 88.50 \\

  TBPS-CLIP~\cite{empirical} &  & AAAI'24 & 61.95 & 83.55 & 88.75 \\

  \hline\hline
  \textbf{Ours} &  & - & \textbf{61.87}  & \textbf{83.63}  & \textbf{89.70}   \\
  \hline
\end{tabular}}
\end{table}

\begin{table*}[!ht]\small
\setlength{\abovecaptionskip}{0.1cm} 
\setlength{\belowcaptionskip}{-0.2cm}
\centering {\caption{Ablation study on different components of our Propot on CUHK-PEDES.}\label{Tab:4}
\renewcommand\arraystretch{1.0}
\begin{tabular}  
{m{0.5cm}<{\centering}|m{2.8cm}<{\centering}|m{0.7cm}<{\centering}|m{0.7cm}<{\centering}|m{0.7cm}<{\centering}|m{0.7cm}<{\centering}|m{0.7cm}<{\centering}|m{0.7cm}<{\centering}m{1.0cm}<{\centering}m{1.0cm}<{\centering}|m{1.0cm}<{\centering}m{1.0cm}<{\centering}}
\hline
\hline
\multirow{2}{*}{No.} & \multirow{2}{*}{Methods} & \multirow{2}{*}{IniPt} &\multirow{2}{*}{DPP} & \multicolumn{2}{c|}{IPP}  & \multirow{2}{*}{MLM} & \multirow{2}{*}{R@1} & \multirow{2}{*}{R@5} & \multirow{2}{*}{R@10} & \multirow{2}{*}{Params} & \multirow{2}{*}{FLOPs}\\
\cline{5-6}
          &  &  &  & Intra & Inter  &  &  &       &       &  & \\
\hline
0\# & Baseline  &  &   &  &  &  & 70.06 & 86.91 & 92.01 & 155.26M  & 20.266 \\

2\# & +IniP  &\checkmark  &  &  &    &  & 73.08 & 88.97 & 93.19 & 155.26M  & 20.278  \\

3\# & +DPP  &\checkmark  &\checkmark  &  &    &  &  74.11 & 89.46 & 93.57 & 203.48M  & 25.704  \\

4\# & +IPP (Intra)  &\checkmark &  &\checkmark   &  &  & 73.83 & 89.21 & 93.53  & 158.41M  & 23.058  \\

5\# & +IPP (Inter)  &\checkmark &  &   &\checkmark  &  & 73.65 & 89.42 & 93.69  & 158.41M  & 23.058  \\

6\# & +IPP  &\checkmark &  &\checkmark   &\checkmark  &  & 74.03 & 89.47 & 93.66  & 158.41M  & 25.838  \\

7\# & +DPP+IPP   &\checkmark &\checkmark &\checkmark  &\checkmark  &  & 74.37 & 89.59 & 93.88 & 206.64M  & 31.264  \\

8\# & +DPP+IPP+MLM  &\checkmark &\checkmark  &\checkmark  &\checkmark  &\checkmark & 74.89 & 89.90 & 94.17 & 245.91M  & 37.353  \\
\hline\hline
\end{tabular}}
\end{table*}

\textbf{Evaluation on ICFG-PEDES:}
Table~\ref{Tab:2} reports the comparative results on the ICFG-PEDES dataset. The findings demonstrate that our Propot establishes a new state-of-the-art performance on the ICFG-PEDES dataset, achieving R@1, R@5, and R@10 accuracy scores of 65.12\%, 81.57\%, and 86.97\%, respectively. Notably, Propot significantly outperforms the recent method~\cite{mgcn} based on single-modal pre-training by +4.92\%, +4.82\%, and +3.07\% on R@1, R@5, and R@10, highlighting the substantial impact of multi-modal prior information in the TIReID task. Moreover, Propot surpasses the current state-of-the-art solution TBPS-CLIP~\cite{empirical} by 1.23\% in R@5 and 1.50\% in R@10. The superior performance on ICFG-PEDES confirms the superiority and generalization of our Propot.

\textbf{Evaluation on RSTPReid:}
We finally evaluate the effectiveness of our method on the latest RSTPReid dataset. 
The comparative analysis with state-of-the-art methods is summarized in Table~\ref{Tab:3}. Propot demonstrates commendable performance, achieving competitive results over recent state-of-the-art methods, specifically attaining 61.83\%, 83.45\%, and 89.70\% on R@1, R@5, and R@10. Specifically, our method achieves suboptimal results, with performance slightly lower (-0.08\%) than the optimal method TBPS-CLIP~\cite{empirical}. TBPS-CLIP represents an empirical study incorporating CLIP into the TIReID task, utilizing various data augmentation and training tricks. In contrast, our approach uses only basic data augmentation without additional tricks. Propot still achieves superior performance on the RSTPReid dataset with complex scenes, underscoring the efficacy of our prototype learning method in seamlessly integrating diverse identity information for modeling identity-level matching. This further establishes the generalization and robustness of Propot.

\subsection{Ablation Studies}
We conduct ablation experiments for Propot on CUHK-PEDES using the default settings above. 
In addition, Baseline solely includes visual and textual encoders initialized by CLIP, which is trained using SDM and cross-entropy loss, with the training settings aligned with those of Propot.

\textbf{Contributions of Proposed Components:}
In Table~\ref{Tab:4}, we assess the contribution of each module of Propot, including the initial prototype (IniPt) generated by CLIP, the DPP module, the IPP module, and the MLM module. Baseline achieves a notable R@1 accuracy of 70.06\% due to the rich multi-modal prior information of CLIP. The introduction of our prototype learning process based on Baseline, aimed at modeling identity-level matching, yields several key observations. Firstly, utilizing only the initial prototype generated by CLIP to supervise identity-level matching leads to significant performance improvements (+3.02\% R@1 improvement over Baseline), affirming the feasibility of our approach and the importance of identity-level matching. Secondly, the incorporation of the DPP module to update the initial prototype results in a 4.05\% R@1 accuracy improvement over Baseline, demonstrating the effective adaptation of the initial prototype to the TIReID task. Thirdly, when the IPP module is employed to update the prototype, whether conditioned on intra-modal or inter-modal instances, substantial performance enhancements are observed (+3.77\% or +3.59\% R@1 improvement over Baseline). The performance is further elevated when both IPP modules are utilized concurrently, underscoring the IPP module's capacity to enrich the prototype with instance information, ensuring its diversity. Fourthly, the collaborative use of DPP and IPP modules further enhances the R@1 accuracy to 74.37\%, surpassing most state-of-the-art methods in Table~\ref{Tab:1}. This is attributed to our Propot's ability to model instance-level and identity-level matching simultaneously. Finally, the incorporation of the local matching module culminates in the best performance for Propot, achieving a remarkable 74.89\% R@1.

\begin{figure}[!t]
\setlength{\abovecaptionskip}{0.1cm} 
\setlength{\belowcaptionskip}{-0.2cm}
\subfigbottomskip=-2pt
\subfigcapskip=-2pt
\centering
\subfigure[impact of $K$] {\includegraphics[height=1.2in,width=1.65in,angle=0]{images/ctx_num.pdf}}
\subfigure[impact of $N_a$] {\includegraphics[height=1.2in,width=1.65in,angle=0]{images/lptlayers.pdf}}
\subfigure[impact of $N_e$] {\includegraphics[height=1.2in,width=1.65in,angle=0]{images/isplayers.pdf}}
\subfigure[impact of $\lambda_1$] {\includegraphics[height=1.2in,width=1.65in,angle=0]{images/plossw.pdf}}
\caption{Effects of four hyper-parameters on CUHK-PEDES, including contextual vector length $K$, the block number $N_a, N_e$, and loss weight $\lambda_1$.}
\label{Fig:3}
\end{figure}

\textbf{Impact of contextual vector length $K$ in DPP:}
To facilitate the adaptation of the initial prototype to the TIReID task, we introduce a set of learnable contextual prompt vectors for each prototype in the DPP module. The length $K$ of this set of vectors is a crucial parameter influencing the prototype's adaptation. To investigate the impact of different vector lengths, we conduct several experiments by varying the value of the context vector length from 1 to 6 and report the results in Figure~\ref{Fig:3} (a). The observed trend indicates that larger values of vector length result in superior performance, as they provide sufficient contextual parameters to extract TIReID task information. Notably, when $K$ equals 4, the performance reaches a peak of 74.89\%. However, excessively large vector lengths may introduce information redundancy, leading to overfitting and a surge in computational costs. Therefore, we set $K$ to 4 in our experiments to strike a balance between performance and efficiency.

\textbf{Influence of $N_a, N_e$:}
The parameters $N_a$ and $N_e$ correspond to the number of blocks of SAE and CAD in DPP and IPP modules, respectively, and their impact on performance is shown in Figure~\ref{Fig:3} (b) and (c). Regarding $N_a$, it is evident that when its value exceeds 3, the performance experiences a significant drop. This suggests that an excessive number of SAE parameters might hinder the learning of contextual prompt vectors. Therefore, we set $N_a$ to 1 in our experiments, which yields the optimal result. In comparison, our model demonstrates less sensitivity to $N_e$, and its curve exhibits a relatively flat trend. Nevertheless, a $N_e$ that is excessively large would introduce a surplus of parameters, leading to a considerable increase in computational costs. Consequently, we set $N_e$ to 3 in our experiments, achieving superior performance while maintaining efficiency.

\begin{table}[!t]\small
\setlength{\abovecaptionskip}{0.1cm} 
\setlength{\belowcaptionskip}{-0.2cm}
\centering {\caption{Ablation study of prototype aggregation schemes on CUHK-PEDES.}\label{Tab:5}
\renewcommand\arraystretch{1.0}
\begin{tabular}{m{3cm}<{\centering}|m{1.2cm}<{\centering}m{1.2cm}<{\centering}m{1.2cm}<{\centering}}
 \hline
  Method  & Rank-1 & Rank-5 & Rank-10 \\
  \hline
  Sum      & 74.30  &  89.86 & 93.84   \\
  
  Average   & 74.29 & 89.39 & 93.34  \\

  MLP     & 73.51 & 89.56 & 93.76 \\

  Parameter   & 74.19 & 89.10 & 93.18  \\
  \hline\hline
  APA (Ours)   & 74.89 & 89.90 & 94.17  \\
  \hline
\end{tabular}}
\end{table}

\textbf{Impact of weight $\lambda_1$ in the objective function:}
The parameter $\lambda_1$ represents the weight of the prototype-to-instance contrastive loss, governing the strength of identity-level matching. To assess the impact of this parameter, we conduct several experiments, varying the value of $\lambda_1$ from 0.01 to 10, with results depicted in Figure~\ref{Fig:3} (d). It is evident from the results that as $\lambda_1$ increases, the performance gradually improves, reaching its peak when $\lambda_1$ is set to 0.2. Subsequently, performance begins to decline as $\lambda_1$ continues to increase, ultimately collapsing when $\lambda_1$ equals 10. This behavior is attributed to the fact that a very small $\lambda_1$ fails to ensure the effective diffusion of identity information to each instance, rendering identity-level matching ineffective. Conversely, an excessively large $\lambda_1$ results in the over-diffusion of identity information, disrupting instance-level matching. Therefore, we set $\lambda_1$ to 0.2 to strike a balance between instance-level and identity-level matching.

\begin{figure*}[!t]
  \centering
  \setlength{\abovecaptionskip}{0.1cm} 
  \setlength{\belowcaptionskip}{-0.2cm}
  \includegraphics[width=\linewidth]{images/retrieval-result.pdf}\\
  \caption{Retrieval result comparisons of Baseline and our Propot on CUHK-PEDES. The matched and mismatched person images are marked with green and red rectangles, respectively.}
  \label{Fig:4}
\end{figure*}

\textbf{Different Prototype Aggregation Schemes:}
To adaptively aggregate multiple prototypes generated by different modules, we devised an Adaptive Prototype Aggregation (APA) module. To demonstrate the superiority of the APA module, we conducted a comparison with several common aggregation schemes, including (1) summing multiple prototypes; (2) averaging multiple prototypes; (3) employing a multi-layer perceptron (MLP) network to generate individual weights for each prototype; (4) parameterizing the weights of each prototype to learn simultaneously with the network. The comparison results are summarized in Table~\ref{Tab:5}. Analysis of the table leads to the following conclusions. Aggregation schemes that introduce parameters to learn aggregation weights perform less favorably than non-parametric aggregation schemes. This discrepancy arises because the introduction of additional parameters amplifies the optimization difficulty and uncertainty of the network. Furthermore, superior performance can be achieved with simple summation or average aggregation schemes, underscoring the feasibility and effectiveness of our idea. The excellence of our aggregation scheme is attributed to its utilization of the initial prototype generated by CLIP as a baseline for adaptive aggregation of other prototypes. The quality of the initial prototype is assured by the powerful semantic understanding capabilities of CLIP.

\textbf{Qualitative Results:}
To further highlight the effectiveness of the proposed Propot, we present some representative retrieval results in Figure~\ref{Fig:4}. For each text description in the query set, Figure 4 displays the top 10 gallery images retrieved by both the baseline and our Propot. The green and red rectangles denote correct and incorrect retrieval results for the given query text, respectively. Baseline solely focuses on instance-level matching between modalities, while our Propot incorporates prototype prompting based on Baseline to model identity-level matching. The examples illustrate that our Propot excels in handling challenging scenarios where the baseline struggles, ensuring that pedestrian images with the same identity as the given query text are ranked highly. This visually underscores the significance of identity-level matching, with our method demonstrating superior performance by concurrently modeling both instance-level and identity-level matching.

\section{Conclusion}
In this study, to model identity-level matching for TIReID, we present Propot, a conceptually simple framework for identity-enriched prototype learning. The framework follows the ‘initialize, adapt, enrich, then aggregate’ pipeline. Initially, we generate robust initial prototypes leveraging the capabilities of CLIP. Subsequently, the Domain-conditional Prototypical Prompting (DPP) module is introduced to prompt and adapt the initial prototypes to the specificities of the TIReID task. To ensure prototype diversity, the Instance-conditional Prototypical Prompting (DPP) is devised, enriching the prototypes through both intra-modal and inter-modal instances. Lastly, an adaptive prototype aggregation module is employed to effectively aggregate multiple prototypes, followed by the diffusion of their rich identity information to instances, thereby facilitating the modeling of identity-level matching. Through extensive ablations and comparative experiments conducted on three widely used benchmarks, we demonstrate the superiority and effectiveness of the proposed Propot framework.

\begin{acks}
This work was supported by...
\end{acks}

\clearpage
\bibliographystyle{ACM-Reference-Format}
\bibliography{sample-base}
